\documentclass{article}

\usepackage[margin = 1in]{geometry}

\usepackage{amsthm,amsmath,amsfonts,amssymb}
\usepackage{graphicx}

\usepackage{setspace}

\usepackage[table]{xcolor}

\makeatletter
\newcommand{\groupedRowColors}[5][0]{% [#1: offset], #2: group size, #3: start line, #4: color 1, #5: color 2
    \global\rownum=\z@
    \global\@rowcolorstrue
    \@ifxempty{#4}%
        {\def\@oddrowcolor{\@norowcolor}}%
        {\def\@oddrowcolor{\gdef\CT@row@color{\CT@color{#4}}}}%
    \@ifxempty{#5}%
        {\def\@evenrowcolor{\@norowcolor}}%
        {\def\@evenrowcolor{\gdef\CT@row@color{\CT@color{#5}}}}%
    \def\@rowcolors{%
        \if@rowcolors
            \noalign{%
                \relax
                \ifnum\rownum<#3
                    \@norowcolor
                \else \ifodd \numexpr (\rownum-#1)/#2\relax
                    \@oddrowcolor
                \else
                    \@evenrowcolor
                \fi \fi
            }%
        \fi
    }%
    \CT@everycr{\@rowc@lors\the\everycr}%
    \ignorespaces
}
\makeatother

\usepackage{lscape}
\usepackage{booktabs}
\usepackage{latexsym}

\usepackage{pifont}
\newcommand{\cmark}{\ding{51}}
\newcommand{\xmark}{\ding{55}}
\usepackage[utf8]{inputenc}
\usepackage{cancel}

\usepackage{longtable}
\usepackage{color, colortbl}

\usepackage{natbib}

\usepackage{algorithm}
\usepackage{algpseudocode} 

\usepackage{hyperref} 

\title{Deep Learning of Semi-Competing Risk Data via a New Neural Expectation-Maximization Algorithm}

\author{
Stephen Salerno \\
Public Health Sciences Division, Biostatistics, Fred Hutchinson Cancer Center 
\and
Zhilin Zhang \\
Department of Biostatistics, University of Michigan
\and
Yi Li \\
Department of Biostatistics, University of Michigan
}

\date{}

\begin{document}

\maketitle

\begin{abstract}

Prognostication for lung cancer, a leading cause of mortality, remains a complex task, as it needs to quantify the associations of risk factors and health events spanning a patient's entire life. One challenge is that an individual's disease course involves non-terminal (e.g., disease progression) and terminal (e.g., death) events, which form {\it semi-competing} relationships. Our motivation comes from the Boston Lung Cancer Study, a large lung cancer survival cohort, which investigates how risk factors influence a patient's disease trajectory. Following developments in the prediction of time-to-event outcomes with neural networks, deep learning has become a focal area for the development of risk prediction methods in survival analysis. However, limited work has been done to predict multi-state or semi-competing risk outcomes, where a patient may experience adverse events such as  disease progression prior to death. We propose a novel neural expectation-maximization algorithm to bridge the gap between classical statistical approaches and machine learning. Our algorithm enables estimation of the nonparametric baseline hazards of each state transition, risk functions of predictors, and the degree of dependence among different transitions, via a multi-task deep neural network with transition-specific sub-architectures. We apply our method to the Boston Lung Cancer Study and investigate the impact of clinical and genetic predictors on disease progression and mortality.

\end{abstract}

%=== MAIN TEXT ==================================================================

%--- SECTION 1 ------------------------------------------------------------------

\section{Background}
\label{sec:1}

Lung cancer remains one of the leading cause of cancer mortality, with a 5-year survival rate of less than 20\% worldwide \citep{bade2020lung, ries2006seer}. Accurately predicting lung cancer prognosis is challenging, as it often depends on risk factors and health events spanning a patient's entire life \citep{le2018application, goel2021correlates}. Moreover, an individual's disease course often involves non-terminal  health events, such as disease recurrence and progression, which form {\it semi-competing} relationships with mortality. By semi-competing, we mean that death may dependently censor disease recurrence or progression, whereas the reverse is not true \citep{fine2001semi}. Failing to account for this relationship may lead to biased predictions and incorrect conclusions \citep{jazic2016beyond}. Further, prognosis varies greatly for patients with lung cancer, and accurate prediction of long-term events such as progression or mortality depends on individualized risk factors including smoking status, genetic variants, and comorbidities \citep{brundage2002prognostic, ashworth2014individual, gaspar2012small}. More broadly, semi-competing risk relationships arise in other chronic conditions such as dialysis, where fluid overload strongly affects both intermediate outcomes and mortality \citep{shu2018effect}.

Our work is motivated by the Boston Lung Cancer Study (BLCS), one of the largest lung cancer survival cohorts in the world \citep{christiani2017blcs}. A primary objective of the BLCS is to better understand how risk factors and adverse events such as disease progression influence a patient's survival. Furthermore, it is important to quantify the dependence between disease progression and mortality in cancer research, as disease progression can potentially act as a precursor to death \citep{inamura2010lung}. The BLCS has amassed a comprehensive database of patients enrolled at the Massachusetts General Hospital and the Dana-Farber Cancer Institute, with data on demographics, social history, pathology, treatments, oncogenic mutation status, and other pertinent risk factors \citep{lynch2004activating, paez2004egfr}.

\textcolor{black}{Deep learning has sparked interest in the advancement of risk prediction methods within the field of survival analysis \citep{faraggi1995neural, katzman2016deep, ranganath2016deep, jing2017neural, kvamme2019time, hao2021deep}. Many of these approaches extend the Cox proportional hazards model \citep{cox1972regression} to nonlinear predictions or use a patient's survival status directly as a binary training label, predicting a patient's survival probability.} More recently, competing risk and multi-state models extend these methods to settings where multiple event types mutually censor one another \citep{lee2018deephit, lee2019dynamic, aastha2020deepcompete, tjandra2021hierarchical}. Such methods characterize the risk of one or more competing events by estimating either the cause-specific or subdistribution hazards of each event type. \textcolor{black}{However, these do not accommodate the joint prediction of correlated events or the study of outcome trajectories between events, important considerations for semi-competing risks.}

\textcolor{black}{Several recent works have developed methods for risk prediction with semi-competing outcomes. \cite{xu2010statistical} proposed an approach based on the illness-death model, which was defined by the hazards of transitioning between disease states. The authors used a shared gamma frailty conditional Markov model, parameterized by three Cox-based hazard functions, and a semiparametric maximum likelihood estimation (MLE) approach. The gamma frailty, a type of subject-specific random effect, captures the strength of unobserved, individual-level heterogeneity that drives dependence between the illness and death transitions \citep{li2020shared}. Estimating its variance allows us to quantify latent risk subgroups, assess how much of the observed transition dynamics are driven by shared frailties versus measured covariates, and thus directly informs prognostic stratification and personalized treatment planning. \cite{lee2015bayesian} also adopted the gamma frailty formulation of the illness-death model with Cox-type hazards, but instead proposed a semiparametric Bayesian approach for estimation. \cite{lee2017accelerated} formulated Bayesian (semi)parametric approaches with an illness-death accelerated failure time (AFT) model, adopting an additive normal frailty, rather than a multiplicative gamma frailty.}

\textcolor{black}{More recently, \cite{gorfine2021marginalized} developed a Cox-based marginalized gamma frailty illness-death model and estimated it using a semiparametric pseudo-likelihood, and \cite{kats2022accelerated} proposed an AFT-based gamma frailty model via semiparametric MLE. In addition, approaches such as the one proposed by \cite{jiang2017semi} consider transformation illness-death models with parametric error distributions, but nonparametric frailty distributions. In high dimensions, \cite{reeder2022penalized} proposed a regularized estimation approach which combines a non-convex and structured fusion penalization. \cite{salerno2022high} developed a deep learning framework for predicting semi-competing risk outcomes based on the model of \cite{xu2010statistical}, with parametric baseline hazards estimated via gradient methods. However, it remains challenging to estimate nonparametric baseline hazards, which confer greater robustness, within a deep learning framework.}

To address this, we introduce a novel neural expectation-maximization (NEM) algorithm, which bridges the gap between classical statistical approaches and machine learning. \textcolor{black}{Our proposal is to replace the traditional parametric M-step with two non-parametric updates. In our M-step, we exploit closed-form non-parametric maximum likelihood estimates to recover the baseline hazards of transitioning between model states. In our N-step, we fit a multi-head neural network via stochastic gradient descent to learn flexible covariate-dependent risk functions. Because the semi-competing illness-death model's likelihood decomposes additively across state transitions, each network head optimizes a transition-specific loss function. The NEM algorithm enables estimation of nonparametric baseline hazards, nonparametric risk functions of our predictors, and the degree of subject-specific dependence between transitions. As deep learning can recover nonlinear risk scores, we assess our method by simulating risk surfaces of varying complexity. We apply our method to the BLCS to investigate the impacts of clinical  and genetic predictors on disease progression and mortality.}

In the following, Section \ref{sec:2} reviews the illness-death model, a compartment-type model for studying the hazards, or transition rates, between semi-competing events. Section \ref{sec:3} proposes a novel neural expectation-maximization algorithm and outlines our approach for semi-competing risk prediction. \textcolor{black}{Section \ref{sec:4} introduces a new framework for evaluating predictive performance in this setting by extending the widely-used Brier score and concordance index for censored univariate time-to-event data to the bivariate survival function.} We  assess the predictive accuracy of our method in Section \ref{sec:5}. We apply our method to analyze the BLCS cohort in Section \ref{sec:6}. We conclude with a discussion and directions for future work.

%--- SECTION 2 ------------------------------------------------------------------

\section{Notation}
\label{sec:2}

\begin{figure}[!ht]
    \centering
    \includegraphics[width=4.125in]{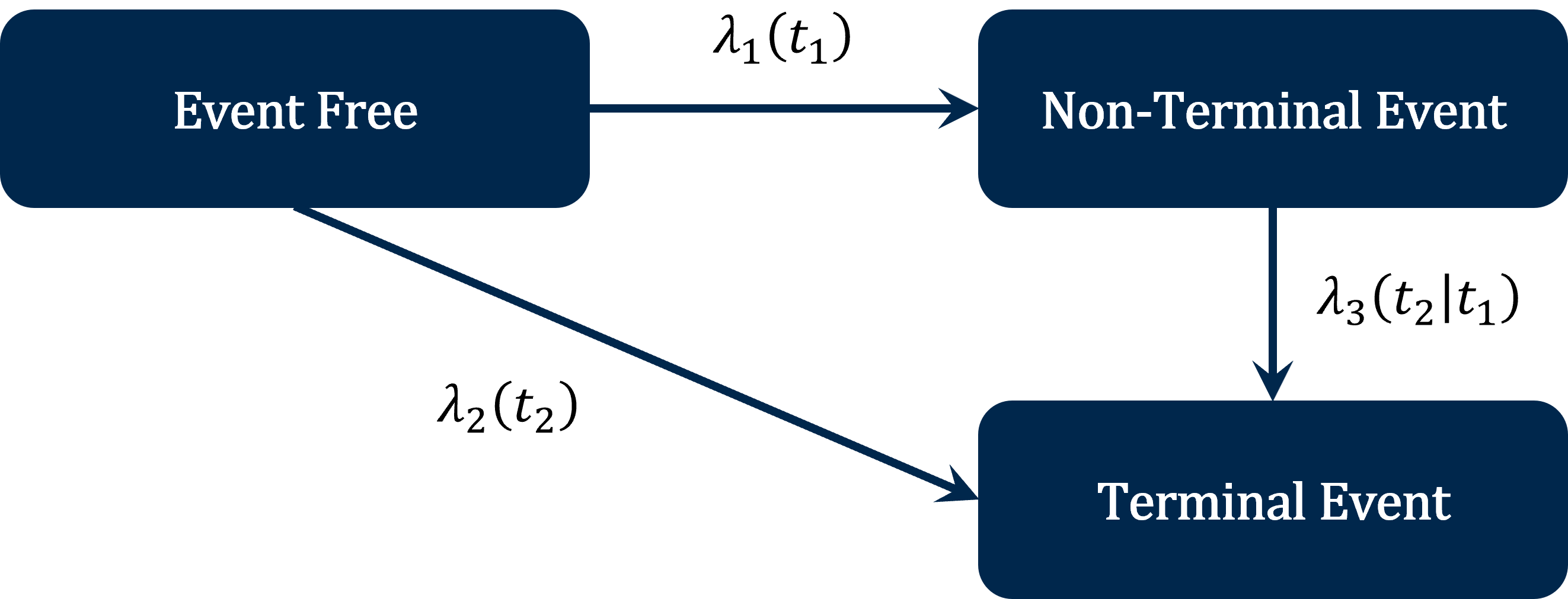}
    \caption{Graphical representation of the illness-death model with three states. Transition rates between states are characterized by $\lambda_1(t_1)$, $\lambda_2(t_2)$, and $\lambda_3(t_2 \mid t_1)$, respectively.}
    \label{fig:1}
\end{figure}

Denote \textcolor{black}{by $T_1$ and $T_2$ the times from an initial, event-free state (e.g., time of diagnosis) to a non-terminal (i.e., recurrence or progression) or terminal (i.e., death) state, respectively, where observing the non-terminal event is subject to the terminal event.} Within the framework of illness-death models \citep{andersen2012statistical}, we  model  the rates at which individuals transition between three states: an event-free state, a non-terminal event state, and a terminal event state. \textcolor{black}{As seen in Figure \ref{fig:1}, $\lambda_1(t_1)$  corresponds to the transition hazard from event-free to the non-terminal event at $t_1 > 0$; $\lambda_2(t_2)$ is the transition hazard from event-free to the terminal event at $t_2 > 0$, given that no non-terminal event had happened by $t_2$, whereas $\lambda_3(t_2 \mid t_1)$ is the transition hazard to the terminal event at $t_2 > 0$, given that the non-terminal event occurred at $t_1 (< t_2)$. These hazards are specified as}

\textcolor{black}{
\begin{align}
    \lambda_1(t_1) &= \lim_{\Delta \rightarrow 0} \Pr[T_1 \in [t_1, t_1 + \Delta) \mid T_1 \geq t_1, T_2 \geq t_1] / \Delta, \label{eq:lam1} \\
    \lambda_2(t_2) &= \lim_{\Delta \rightarrow 0} \Pr[T_2 \in [t_2, t_2 + \Delta) \mid T_1 \geq t_2, T_2 \geq t_2] / \Delta, \label{eq:lam2} \\
    \lambda_3(t_2 \mid t_1) &= \begin{cases} \lim_{\Delta \rightarrow 0} \Pr[T_2 \in [t_2, t_2 + \Delta) \mid T_1 = t_1, T_2 \geq t_2] / \Delta, & t_2 > t_1 > 0; \\ 0, & {\rm otherwise}. \end{cases} \label{eq:lam3}
\end{align}
}

\begin{figure}[!ht]
    \centering
    \includegraphics[width = 2.75in]{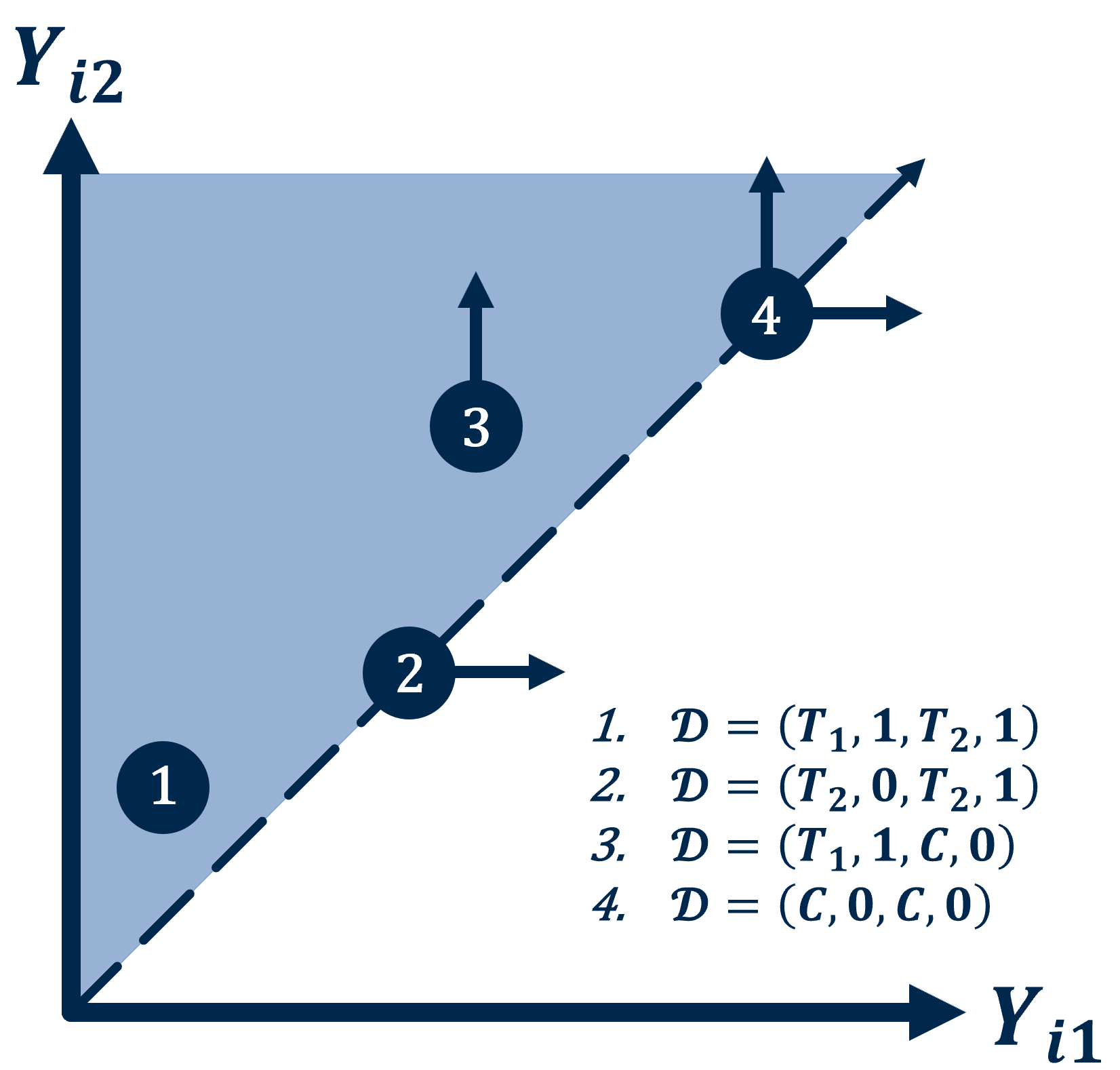}
    \caption{Graphical representation of the observable space for semi-competing data with example observations: (1) both events are observed, (2) only the terminal event is observed, (3) only the non-terminal event is observed, and (4) neither event is observed. The arrows represent the direction of censoring, and $\mathcal{D} = (Y_1, \delta_1, Y_2, \delta_2)$ represents the data under each example observation.}
    \label{fig:2}
\end{figure}

\textcolor{black}{Similar formulations have been proposed by \cite{xu2010statistical, gorfine2021marginalized, kats2022accelerated}, and others.} Both the non-terminal and terminal events can be subject to independent censoring; we focus only on the case of right censoring, whereby a subject may be lost to follow-up or the study ends before the event has occurred. For the $i$th individual in a sample of $n$ subjects, we denote the censoring time by $C_i$ and add a subscript $i$ to $T_1, T_2$ for this individual. The observed data are denoted as
\[
\mathcal{D} = \{(Y_{i1},\ \delta_{i1},\ Y_{i2},\ \delta_{i2});\ i=1,\ldots,n\},
\]

\noindent where $Y_{i2} = \min(T_{i2}, C_i)$, $\delta_{i2} = I(T_{i2} \leq C_i)$, $Y_{i1} = \min(T_{i1}, Y_{i2})$, $\delta_{i1} = I(T_{i1} \leq Y_{i2})$, and $I(\cdot)$ denotes the indicator function. Note that our observable data take on probability mass only in the {\it upper wedge} on which $Y_{i1} \leq Y_{i2}$ and arise from four potential cases{: (1) a subject experiences both event types, (2) a subject experiences only the terminal event, (3) a subject experiences only the non-terminal event, or (4) a subject experiences neither event prior to the end of follow up} (Figure \ref{fig:2}). \textcolor{black}{Following the original work of \cite{xu2010statistical}, we model (\ref{eq:lam1}) - (\ref{eq:lam3}) by extending a Cox-type hazard function for each state transition to include a baseline hazard, a frailty term, and a patient's covariates as}

\textcolor{black}{
\begin{align}
    \lambda_1(t_1 \mid \gamma_i, \boldsymbol{x}_i) &= \gamma_i \lambda_{01}(t_1) \exp\{h_1(\boldsymbol{x}_i)\}, \label{eq:condhaz1} \\
    \lambda_2(t_2 \mid \gamma_i, \boldsymbol{x}_i) &= \gamma_i \lambda_{02}(t_2) \exp\{h_2(\boldsymbol{x}_i)\}, \label{eq:condhaz2} \\
    \lambda_3(t_2 \mid t_1, \gamma_i, \boldsymbol{x}_i) &= \begin{cases} \gamma_i \lambda_{03}(t_2) \exp\{h_3(\boldsymbol{x}_i)\}, & t_2 > t_1 > 0; \\ 0, & {\rm otherwise}, \end{cases} \label{eq:condhaz3}
\end{align}
}

\noindent \textcolor{black}{where $\gamma_{i}$ is a subject-specific random effect, termed {\it frailty}, that induces dependence among the three transition processes, $\lambda_{01}\left(t_{1}\right)$, $\lambda_{02}\left(t_{2}\right)$, and $\lambda_{03}(t_{2})$ are the baseline hazard functions for the three state transitions, respectively, $\boldsymbol{x}_{i}$ is a $p$-vector of clinically relevant, time-independent predictors such as patient socio-demographic status, medical history data, and comorbid conditions collected at baseline, and $h_g(\boldsymbol{x}_i), g\in\{1,2,3\},$ are log-risk functions which relate a patient's covariates to the hazard rates for each potential transition.} Unlike existing works, we do not parameterize the $h_g(\boldsymbol{x}_i)$. Instead, we estimate these functions nonparametrically as outputs from our proposed neural network architectures. \textcolor{black}{For identifiability, we constrain the bias terms in the output layer to zero and assume that each $\gamma_i$ are Gamma distributed with $\mathbb{E}(\gamma_i) = 1$ and $\text{Var}(\gamma_i) = \theta$. The Gamma frailty is a common assumption in literature \citep{xu2010statistical, haneuse2016semi, kats2022accelerated}, however it is unverifiable in the bivariate distribution's lower wedge (see Figure \ref{fig:2}). This limitation is not specific to this work, rather a feature of the semi-competing risks paradigm in the absence of even stronger unverifiable assumptions. Alternatives such as finite mixtures have recently been suggested as well \citep{gasperoni2020non, chee2021semiparametric}.  We employ the frailty model in our study, as investigating this dependence is one of the objectives of the motivating BLCS study. If, however, the dependence parameter $\theta$ was considered to be a nuisance, an alternative approach would be the marginalized model proposed by \cite{gorfine2021marginalized}. Denoting the vector of frailties by $\boldsymbol{\gamma} = (\gamma_1, \ldots, \gamma_n)$, the cumulative baseline hazards for each transition by $\Lambda_{0g}(t) = \int_0^t \lambda_{0g}(u)du$ for $g= 1,2,3$, and the collection of model parameters by $\boldsymbol{\psi} = \{\Lambda_{01}, \Lambda_{02}, \Lambda_{03}, h_1, h_2, h_3, \theta\}$, the `complete' data likelihood is}

\textcolor{black}{
\begin{align}
\begin{split}
\label{eq:condlik}
    L &\left(\boldsymbol{\psi}; \mathcal{D}, \boldsymbol{\gamma}\right) = \prod_{i=1}^{n} \frac{\theta^{-\frac{1}{\theta}}}{\Gamma\left(\frac{1}{\theta}\right)} \times \gamma_{i}^{\frac{1}{\theta}-1} \times e^{-\frac{\gamma_{i}}{\theta}}\times \gamma_{i}^{\delta_{i 1}+\delta_{i 2}} \times \left[\lambda_{01}\left(Y_{i1}\right) e^{h_1(\boldsymbol{x}_i)}\right]^{\delta_{i 1}} \\[1ex]
    & \times \left[\lambda_{02}\left(Y_{i2}\right)e^{h_2(\boldsymbol{x}_i)}\right]^{\left(1-\delta_{i 1}\right) \delta_{i 2}} \times \left[\lambda_{03}\left(Y_{i2} \right)e^{h_3(\boldsymbol{x}_i)}\right]^{\delta_{i 1} \delta_{i 2}}\\[1ex]
    &\times \exp\left\{-\gamma_{i}\left[\Lambda_{01}\left(Y_{i1}\right)e^{h_1(\boldsymbol{x}_i)} + \Lambda_{02}\left(Y_{i1}\right)e^{h_2(\boldsymbol{x}_i)}\right.\right.\\
    &\quad\quad \left.\left.+\ \delta_{i 1}[\Lambda_{03}(Y_{i2}) - \Lambda_{03}(Y_{i1})] e^{h_3(\boldsymbol{x}_i)}\right]\right\}.
\end{split}
\end{align}
}

\vspace{1ex}

\noindent \textcolor{black}{See Supplement A for the derivation of  (\ref{eq:condlik}). One could further integrate out $\gamma_{i}$ in (\ref{eq:condlik}) and maximize the integrated likelihood with respect to the unknown parameters. However, two challenges arise. The first challenge is the lack of a closed-form for the integral, making computation difficult. The second challenge stems from the non-existence of a maximizer for the function over the space of absolutely continuous cumulative baseline hazards \citep{johansen1983extension}. To address the first challenge, we introduce an EM-type algorithm, treating the $\gamma_i$'s as missing variables and estimating the parameters iteratively. For the second challenge, we constrain the parameter space of cumulative baseline hazards, $\Lambda_{01}$, $\Lambda_{02}$, and $\Lambda_{03}$, to consist of piecewise constant CADLAG (right-continuous with left-hand limits) functions, with jumps at observed event times. The maximizers over this discrete space are termed nonparametric maximum likelihood estimates (NPMLEs) of $\Lambda_{01}$, $\Lambda_{02}$, and $\Lambda_{03}$. In this framework, we modify the likelihood function (\ref{eq:condlik}) by replacing $\lambda_{0g}(t)$ with $\Delta\Lambda_{0g}(t)=\Lambda_{0g}(t)-\Lambda_{0g}(t-)$, representing the jump size at $t$ \citep{li2000covariate, kim2012joint}, and obtain}

\textcolor{black}{
\begin{align}
\begin{split}
\label{eq:condlik2}
    \tilde{L} &\left(\boldsymbol{\psi}; \mathcal{D}, \boldsymbol{\gamma}\right) = \prod_{i=1}^{n} \frac{\theta^{-\frac{1}{\theta}}}{\Gamma\left(\frac{1}{\theta}\right)} \times \gamma_{i}^{\frac{1}{\theta}-1} \times e^{-\frac{\gamma_{i}}{\theta}}\times \gamma_{i}^{\delta_{i 1}+\delta_{i 2}} \times \left[\Delta\Lambda_{01}\left(Y_{i1}\right) e^{h_1(\boldsymbol{x}_i)}\right]^{\delta_{i 1}} \\[1ex]
    & \times \left[\Delta\Lambda_{02}\left(Y_{i2}\right)e^{h_2(\boldsymbol{x}_i)}\right]^{\left(1-\delta_{i 1}\right) \delta_{i 2}} \times \left[\Delta\Lambda_{03}\left(Y_{i2} \right)e^{h_3(\boldsymbol{x}_i)}\right]^{\delta_{i 1} \delta_{i 2}}\\[1ex]
    &\times \exp\left\{-\gamma_{i}\left[\Lambda_{01}\left(Y_{i1}\right)e^{h_1(\boldsymbol{x}_i)} + \Lambda_{02}\left(Y_{i1}\right)e^{h_2(\boldsymbol{x}_i)}\right.\right. \\
    &\quad\quad \left.\left.+\ \delta_{i 1}[\Lambda_{03}(Y_{i2}) - \Lambda_{03}(Y_{i1})] e^{h_3(\boldsymbol{x}_i)}\right]\right\}.
\end{split}
\end{align}
}

\vspace{1ex}

\noindent  \textcolor{black}{This `complete' likelihood (\ref{eq:condlik2}), defined on a step-wise constant space for  $\Lambda_{0g}$, will serve as the basis our proposed neural expectation-maximization algorithm, detailed below.}  

%--- SECTION 3 ------------------------------------------------------------------

\section{Neural Expectation-Maximization Algorithm}
\label{sec:3}

%Classically, to estimate the baseline hazard functions nonparametrically, one can carry out nonparametric maximum likelihood estimation (NPMLE) \citep{kim2012joint}, where the parameter space of $\Lambda_{0g}$ contains non-decreasing step functions with jumps at observed failure times. 
\textcolor{black}{One challenge stands out when we design our EM algorithm, as the risk functions, $h_g$, are completely nonparametric. Therefore, we propose to incorporate a deep learning step into the M-step, which maximizes the expected log conditional likelihood given the data and the current parameter estimates. We term the enhanced algorithm the ``neural expectation-maximization algorithm.'' Specifically, by viewing the frailty term as a missing variable, the algorithm iterates between an expectation (E), a maximization (M) and a neural (N; i.e., deep learning) step. In the E-step, we compute the conditional expectation of the log-likelihood (\ref{eq:condlik2}) given the observed data, $\mathcal{D}$, and the current estimates of $\boldsymbol{\psi}$, denoted by $\boldsymbol{\psi}_c$, wherein the conditional expectation is with respect to the distribution of $\gamma_i \mid \mathcal{D}, \boldsymbol{\psi}_c $. In the M-step, we calculate the NPMLEs of $\Lambda_{0g}$ while assuming the known values of $h_g$. Subsequently, we substitute these estimates into the conditional expectation of the log-likelihood (\ref{eq:condlik2}). This process yields the conditional expectation of the log `profile' likelihood, serving as the objective function for utilizing deep neural networks to derive estimates for the log risk functions and frailty variance.}

\subsection{Conditional Frailty Distribution}
\label{sec:3.1}

It follows that the conditional distribution of $\gamma_i,$  given $\mathcal{D} \, \rm{and} \, \boldsymbol{\psi}$,
is $\text{Gamma}(\tilde{a}, \tilde{b})$, where
\textcolor{black}{
\[
\begin{aligned}
    \tilde{a} &= \frac{1}{\theta} + \delta_{i1} + \delta_{i2}, \\
    \tilde{b} &= \frac{1}{\theta} + \Lambda_{01}\left(Y_{i1}\right) \exp\{h_1(\boldsymbol{x}_i)\} +\ \Lambda_{02}\left(Y_{i1}\right) \exp\{h_2(\boldsymbol{x}_i)\} \\
    &\quad \quad +\ \delta_{i1} \left\{\Lambda_{03}(Y_{i2}) - \Lambda_{03}(Y_{i1})\right\} \exp\{h_3(\boldsymbol{x}_i)\}. 
\end{aligned}
\]
}

\vspace{1ex}

\noindent Then, $\mathbb{E}[\gamma_i|\mathcal{D}, \boldsymbol{\psi}]= \tilde{a}/\tilde{b}$, and  $\mathbb{E}[\log(\gamma_i) | \mathcal{D}, \boldsymbol{\psi}]= \text{digamma}(\tilde{a}) - \log(\tilde{b})$, where $\text{digamma}(\tilde{a}) = \partial\log[\Gamma(\tilde{a})]/\partial\tilde{a}$ and $\Gamma(\cdot)$ is the gamma function. Both quantities, with $\boldsymbol{\psi}$ replaced by $\boldsymbol{\psi}_c$, are needed for the E-Step (see Supplement B).

\subsection{E-Step}
\label{sec:3.2}

The E-step calculates the expected log-conditional likelihood of the `complete' data given the observed data and the current estimate of parameters, i.e., $\mathcal{D}, \boldsymbol{\psi}_c $:

\begin{equation} \label{exp-cond}
Q\left( \boldsymbol{\psi} | \boldsymbol{\psi}_c\right) = \mathbb{E}\left[ \log \tilde{L}\left(\boldsymbol{\psi} ; \mathcal{D}, \boldsymbol{\gamma}\right) \mid \mathcal{D}, \boldsymbol{\psi}_c\right] = Q_{1} + Q_{2} + Q_{3} + Q_{4},
\end{equation}

\noindent where we recall $\boldsymbol{\psi} = (\Lambda_{0g}, h_g, \theta); g=1,2,3$ represents the unknown parameters to be estimated, or updated, and $Q_{1}$, $Q_{2}$, $Q_{3}$, and $Q_{4}$ are the additive pieces of the `$Q$' function, each involving non-overlapping unknown parameters:

\textcolor{black}{
\[
\begin{aligned}
Q_{1} & = \sum_{i=1}^{n}\delta_{i 1} \mathbb{E}[\log(\gamma_i) | \mathcal{D}, \boldsymbol{\psi}_c] +\delta_{i 1}\left\{\log\left[\Delta\Lambda_{01}\left(Y_{i1}\right)\right] + h_1(\boldsymbol{x}_i)\right\}\\
& \quad\quad\quad -\ \mathbb{E}[\gamma_i | \mathcal{D}, \boldsymbol{\psi}_c] \Lambda_{01}\left(Y_{i1}\right)e^{h_1(\boldsymbol{x}_i)}, \\
Q_{2} & = \sum_{i=1}^{n}\delta_{i 2} \mathbb{E}[\log(\gamma_i) | \mathcal{D}, \boldsymbol{\psi}_c] +\left(1-\delta_{i 1}\right) \delta_{i 2} \left\{\log\left[\Delta\Lambda_{02}\left(Y_{i2}\right)\right] + h_2(\boldsymbol{x}_i)\right\}\\
& \quad\quad\quad -\ \mathbb{E}[\gamma_i | \mathcal{D}, \boldsymbol{\psi}_c] \Lambda_{02}\left(Y_{i1}\right)e^{h_2(\boldsymbol{x}_i)},\\
Q_{3} & = \sum_{i=1}^{n}\delta_{i 1} \delta_{i 2} \left\{\log \left[\Delta\Lambda_{03}\left(Y_{i2}\right)\right] + h_3(\boldsymbol{x}_i)\right\} \\
&\quad\quad\quad -\ \mathbb{E}[\gamma_i |  \mathcal{D}, \boldsymbol{\psi}_c] \delta_{i 1}[\Lambda_{03}(Y_{i2})- \Lambda_{03}I(Y_{i1})] e^{h_3(\boldsymbol{x}_i)}, \\
Q_{4} & = \sum_{i=1}^{n}-\frac{1}{\theta} \log (\theta)+\left(\frac{1}{\theta}-1\right) \mathbb{E}[\log(\gamma_i) | \mathcal{D}, \boldsymbol{\psi}_c] -\frac{1}{\theta}\mathbb{E}[\gamma_i | \mathcal{D}, \boldsymbol{\psi}_c]-\log \Gamma\left(\frac{1}{\theta}\right).
\end{aligned}
\]
}

\subsection{M-Step}
\label{sec:3.4}

In the M-step, we maximize (\ref{exp-cond}) with respect to $\boldsymbol{\psi}$ to obtain its updated values. The separability of $Q_1, \ldots, Q_4$ allows us to estimate $\Lambda_{0g}, h_g$ by maximizing the $Q_g; g=1,2,3$, respectively, and estimate $\theta$ by maximizing $Q_4$. \textcolor{black}{As $\Lambda_{0g}, h_g$ are nonparametric, we adopt a profiling approach to facilitate maximization. For each $g=1,2,3$, we maximize $Q_g$ with respect to the jump sizes of $\Lambda_{0g}$, fixing $h_g$. This yields Breslow-type estimates:}

\textcolor{black}{
\[
\begin{aligned}
\widehat{\Delta\Lambda_{01}}(t) &= \frac{\sum_{i=1}^{n} \delta_{i1} I\left[Y_{i1} = t\right]
}{\sum_{i=1}^{n}\mathbb{E}[\gamma_i|\mathcal{D}, \boldsymbol{\psi}_c] I\left[Y_{i1} \geq t\right]\exp\left\{h_1(\boldsymbol{x}_i)\right\}},\\
\widehat{\Delta\Lambda_{02}}(t) &= \frac{\sum_{i=1}^{n}\left(1-\delta_{i1}\right) \delta_{i2}I\left[Y_{i2} = t\right]}{\sum_{i=1}^{n}\mathbb{E}[\gamma_i | \mathcal{D}, \boldsymbol{\psi}_c] I[Y_{i1} \geq t]\exp\left\{h_2(\boldsymbol{x}_i)\right\}},\\
\widehat{\Delta\Lambda_{03}}(t) &= \frac{\sum_{i=1}^{n}\delta_{i1} \delta_{i2}I\left[Y_{i2} = t\right]}{\sum_{i=1}^{n}\mathbb{E}[\gamma_i |  \mathcal{D}, \boldsymbol{\psi}_c] \delta_{i1} \left[I(Y_{i2} \geq t) - I(Y_{i1} \geq t)\right]\exp\left\{h_3(\boldsymbol{x}_i)\right\}}.
\end{aligned}
\]
}

\vspace{2ex}

\noindent See the detailed derivation in the Appendix. 

\subsection{N-Step}
\label{sec:3.3}

\textcolor{black}{From the M-step, we have estimates $\widehat{\Delta\Lambda_{0g}}(t)$ and $\widehat{\Lambda_{0g}}(t) = \sum_{s \le t} \widehat{\Delta\Lambda_{0g}}(s)$ for $g = 1,2,3$. Plugging these estimates into $Q_1$, $Q_2$, and $Q_3$ yields the expected log-{\it profile} likelihood for $h_1, h_2, h_3$, respectively (up to additive constants; see the Appendix). That is, with an added subscript $P$ (for profile), we have that}  
%$\mathbb{E}_{\boldsymbol{\gamma}}\left[\ell\left(\boldsymbol{\psi}; \mathcal{D}, \boldsymbol{\gamma}\right)\mid \mathcal{D}, \boldsymbol{\psi}\right] = 
%$Q_{1, P}= \sum_{i = 1}^{n} \delta_{i1} \mathbb{E}\left[\log(\gamma_i) \mid \mathcal{D}, \boldsymbol{\psi}\right] + \delta_{i2} \mathbb{E}\left[\log(\gamma_i) \mid \mathcal{D}, \boldsymbol{\psi}\right]$

{\textcolor{black}{
\[
\begin{aligned}
Q_{1, P} &= \sum_{i = 1}^{n} \delta_{i1} \left\{h_1(\boldsymbol{x_i}) - \log[\sum_{j=1}^{n}\mathbb{E}\left[\gamma_j\mid\mathcal{D}, \boldsymbol{\psi}_c\right] I\left(Y_{j1} \geq  Y_{i1}\right) \exp\left\{h_1(\boldsymbol{x}_j)\right\}]\right\}, \\
Q_{2, P} &= \sum_{i = 1}^{n}  (1 - \delta_{i1}) \delta_{i2}\left\{h_2(\boldsymbol{x_i}) - \log[\sum_{j = 1}^{n} \mathbb{E}\left[\gamma_j \mid \mathcal{D}, \boldsymbol{\psi}_c\right] I\left(Y_{j2} \geq Y_{i2}\right)\exp\left\{h_2(\boldsymbol{x}_j)\right\}] \right\}, \\
Q_{3, P} &= \sum_{i = 1}^{n} \delta_{i1} \delta_{i2}\left\{h_3(\boldsymbol{x_i}) - \log[\sum_{j = 1}^{n} \mathbb{E}\left[\gamma_j \mid  \mathcal{D}, \boldsymbol{\psi}_c\right] \delta_{j1} I\left[Y_{j2} \geq \max(Y_{i2}, Y_{j1})\right] \exp\left\{h_3(\boldsymbol{x}_j)\right\}]\right\}.
\end{aligned}
\]
}}

\textcolor{black}{Note that these functions resemble the partial likelihood in the Cox setting. In the N-step, we treat each of the three profile objectives, $Q_{g, P},\ g = 1,2,3$, as losses for distinct ``heads'' of a multitask artificial neural network. Concretely, the network takes each subject's covariate vectors, $\boldsymbol{x_i}$, as input and passes them through three subnetworks, which output estimates for the log-risk functions, $\hat{h}_g(\boldsymbol{x}_i)$, respectively.}
%The N-Step estimates the $h_g(\boldsymbol{x}_i)$ {and the frailty variance, $\theta$,}  as outputs from a multi-task deep neural network with three neural network sub-architectures, corresponding to the three ``profile'' objective functions, $Q_{g,P}, g=1,2,3$,  coupled with $Q_4$ (for $\theta$); see Figure \ref{fig:3}. 
\textcolor{black}{Each subnetwork is a fully-connected feed-forward artificial neural network with $L$ layers and $k_{l}$ neurons in the $l$th layer $(l=1, \ldots, L)$. Each layer applies an activation function, $\sigma_{l}: \mathbb{R}^{k_{l+1}} \rightarrow \mathbb{R}^{k_{l+1}}$, to an affine transformation of the data, given by $\mathbf{W}_{l} \boldsymbol{x}+\mathrm{b}_{l}$, so each layer-specific function is given by
\[
f_{l}(\boldsymbol{x})=\sigma_{l}\left(\mathbf{W}_{l} \boldsymbol{x}+\mathrm{b}_{l}\right) \in \mathbb{R}^{k_{l+1}},
\]
where $\mathbf{W}_{l}$ is a $k_{l+1} \times k_{l}$ weight matrix, and $\mathrm{b}_{l}$ is a $k_{l+1}$-dimensional bias vector. The layers are then connected through an $L$-fold composite function,
\[
F_{L}(\cdot) = f_{L} \circ f_{L-1} \circ \cdots \circ f_{1}(\cdot),
\]
where $(g\circ f)(\cdot) = g(f(\cdot))$. We take nonlinear (e.g., $\sigma_{l}(\boldsymbol{x}) = \max (0,\boldsymbol{x})$) activations in the hidden layers and a linear activation in the final layer \citep{yegnanarayana2009artificial}. For identifiability, we require $h_g(\mathbf{0})=0, g=1, 2, 3$, where $\mathbf{0}$ is a $p \times 1$ vector of 0's. Programatically, this translates to us restricting the bias term in the output layer of the subnetworks to be $\mathbf{0}$.} 

\textcolor{black}{Each subnetwork is trained to maximize its respective profile likelihood. For head $g$, we form the per-sample contribution, $Q_{ig,P}$, where $\sum_i Q_{ig,P} = Q_{g,P}$, and optimize the total loss using mini-batch stochastic gradient descent \citep{amari1993backpropagation}, or an adaptive variant such as Adam \citep{kingma2014adam}. At each iteration, we randomly permute the $n$ subjects, split them into batches of size $B$, and for each batch, perform a ``forward pass'' through all layers to compute $\hat{h}_g$, evaluating their individual losses. Automatic differentiation \citep{paszke2017automatic} then computes the gradients, $\nabla_{\boldsymbol{W}_l}$ and $\nabla_{b_l}$. In the ``backward pass,'' we propagate errors from each head through the network's computational graph and update the parameters
\[
\boldsymbol{W}_l \leftarrow \boldsymbol{W}_l - \eta\nabla_{\boldsymbol{W}_l},\quad b_l\leftarrow b_l - \eta\nabla_{b_l},
\]
where $\eta$ is a tunable parameter known as the `learning rate.' Additionally, dropout (i.e., ``turning off'' random neurons during training) and $\ell_2$‐regularization (i.e., penalizing large weights by adding their squared values to the loss) can be incorporated on each hidden layer to prevent overfitting. The number of hidden layers, nodes per layer, dropout fraction, regularization rate, and learning rate are optimized as hyperparameters over a grid of values based on predictive performance. Specifically, we choose three hidden layers, we selected the nodes per layer from 16 to 1024, we set the dropout fraction to be 0.3, and we tuned the learning rate from 0.0001 to 0.05. We repeat these mini-batch updates for a fixed number of epochs, or until the increase in the expected log-profile likelihood falls below a pre-specified tolerance.}

\textcolor{black}{Lastly, note that the frailty variance, $\theta$, contributes to $Q$ solely through $Q_{4}$. As such, to update $\theta$, we maximize $Q_4$ directly by running a small inner Adam loop. At each inner epoch we zero the gradient, compute $Q_4$, backpropagate, and step, with early‐stopping once the change in loss falls below a prespecified threshold. This decouples the frailty update from the network's forward/backward graph, simplifies implementation, and stabilizes estimation.}

\textcolor{black}{ After convergence of the N-step, the network's outputs, $\hat{h}_g$ and $\hat{\theta}$, become the updated estimates of the log‐risk functions and frailty variance. These estimates, and the estimated baseline hazards from the M-step, are then fed back into the E-step to recompute the conditional expectations, and the entire neural EM loop continues until overall convergence.} Lastly, to initialize the neural EM algorithm, we set $\Lambda_{01}$, $\Lambda_{02}$, and $\Lambda_{03}$ to their corresponding Nelson-Aalen estimates. \textcolor{black}{For $h_g$ and $\theta$, we fit a Weibull-Cox semi-competing risks regression (e.g., using the {\tt SemiCompRisks} R package) and extracted the resulting fitted values and frailty variance estimate as our starting values \citep[see][for details]{alvares2019semicomprisks}. For clarity, we summarize this procedure in Algorithm \ref{alg:neural-em-full}. We implement our approach in PyTorch, a popular machine learning library in Python \citep{stevens2020deep}. The Python code to implement our method, as well as the necessary scripts to reproduce our numerical results are available on GitHub at: \href{https://github.com/salernos/SemiCompDNN}{https://github.com/salernos/SemiCompDNN}}.

% \begin{figure}
%     \includegraphics[width = 5.5in]{images/overview.png}
%     \caption{\textcolor{red}{Overview of the neural expectation-maximization algorithm for semi-competing risks.}}
%     \label{fig:3}
% \end{figure}

\begin{algorithm}[!ht]
\caption{\textcolor{black}{Neural Expectation-Maximization Algorithm}}
\label{alg:neural-em-full}
{\color{black}
\begin{algorithmic}[1]
\Require Observed Data: $\mathcal{D}=\{(Y_{i1},\delta_{i1},Y_{i2},\delta_{i2}, \boldsymbol{x}_i)\}_{i=1}^n$, Initial Values: $\theta^{(0)}, \Lambda_{0g}^{(0)}, h_g(\boldsymbol{x}_i),\ g= 1,2,3$, Network Parameters: $\Theta^{(0)} = \{\boldsymbol{W}_l^{(0)},b_l^{(0)}\}$, Learning Rate: $\eta$, Batch Size: $B$, Max Epochs: $E$, Inner‐Loop Epochs: $E_\theta$, Tolerance: $\tau_1$, Convergence Tolerance: $\tau_2$
\State Set iteration $k\gets0$
\Repeat
  \State $k\gets k+1$
  \Statex \quad\textbf{E–Step (Latent Frailties):}
  \For{$i=1$ to $n$}
    \State Compute 
      $\mathbb{E}[\gamma_i\mid\mathcal{D},\Lambda_{0g}^{(k-1)},\theta^{(k-1)},h_g^{(k-1)}]$ and $\mathbb{E}[\log(\gamma_i)\mid\cdot]$
  \EndFor
  \Statex \quad\textbf{M–Step (Baseline Hazards):}
  \For{$g=1$ to $3$}
    \For{each unique event time $t$}
      \State Update jump size $\Delta\Lambda_{0g}^{(k)}(t)$
    \EndFor
    \State $\Lambda_{0g}^{(k)}(t)\gets\sum_{s\le t}\Delta\Lambda_{0g}^{(k)}(s)$
  \EndFor
  \Statex \quad\textbf{N–Step (Log-Risk Functions \& $\theta$):}
  \State Initialize Adam states for $\Theta$; freeze $\{\Lambda_{0g}^{(k)}, \mathbb{E}[\gamma_i\mid \cdot],\mathbb{E}[\log(\gamma_i)\mid \cdot]\}$
  \For{epoch = $1$ to $E$}
    \State Shuffle and form mini‐batches of size $B$
    \For{each batch $\mathcal{B}$}
      \State \textbf{Forward pass:} compute $\hat h_g(\boldsymbol{x}_i)$ for $i\in\mathcal{B},\,g=1,2,3$
      \For{$g=1$ to $3$}
        \State Compute profile loss $Q_{g, P}$          
      \EndFor
      \State $Q \gets Q_1 + Q_2 + Q_3$
      \State \textbf{Backward pass:} compute $\nabla_{\Theta}L$
      \State \textbf{Adam step:} update $\Theta\gets\text{Adam}(\Theta,\nabla_{\Theta}Q)$
    \EndFor
  \EndFor
  \Statex \quad\textbf{Separate $\theta$ Update:}
  \For{inner = $1$ to $E_\theta$}
    \State Compute $Q_4$
    \State Back propagate
    \State Step
    \If{$|\Delta Q_4|\leq\tau_1$}
      \State \textbf{break}
    \EndIf
  \EndFor
  \State $\theta^{(k)}\gets\text{current value of }\theta$
\Until{$\bigl\|\psi^{(k)}-\psi^{(k-1)}\bigr\|\leq\tau_2$}
\Ensure Final estimates $\{\Lambda_{0g}^{(k)},\ h_g^{(k)}(\cdot),\ \theta^{(k)}\}$
\end{algorithmic}
}
\end{algorithm}

%--- SECTION 4 ------------------------------------------------------------------

\section{Measures of Predictive Performance}
\label{sec:4}

\subsection{Bivariate Brier Score}

To assess predictive performance in semi-competing risk settings, we propose a bivariate extension to the inverse probability of censoring weighting (IPCW)-approximated Brier Score \citep{brier1950verification}. Let $S_i(t) = \Pr(T_{i1} > t,\ T_{i2} > t) $ denote the disease-free survival function for individual $i$ at a given, fixed time point $t$. Further, denote an estimate of $S_i(t)$ by $\pi_i(t)$, e.g., based on  (\ref{eq:condhaz1})- (\ref{eq:condhaz3}). If  $S_i(t)$ were known, a bivariate Brier score would simply be the mean squared error, ${\rm MSE}(t) = \frac{1}{n}\sum_{i=1}^n \left[S_i(t) - \pi_i(t)\right]^2$. With unknown $S_i(t)$, we estimate it with our observed data, and in the presence of censoring. Let $G_i(t) = \Pr(C_i > t) > 0$ be the survival function of the censoring distribution for the $i$th individual. We propose a  bivariate Brier score for assessing $\pi_i(t)$ as follows:

\begin{align}
\begin{split}
{\rm B}&{\rm BS}(t) = \frac{\pi_i(t)^2\cdot I\left(Y_{i1} \leq t,\ \delta_{i1} = 1,\ Y_{i1} \leq Y_{i2}\right\}}{G_i(Y_{i1})}\\ 
&+\ \frac{\pi_i(t)^2\cdot I\left(Y_{i1} \leq t,\ Y_{i2} \leq t,\ \delta_{i1} = 0,\ \delta_{i2} = 1,\ Y_{i1} \leq Y_{i2}\right\}}{G_i(Y_{i2})}\\
&+\ \frac{[1 - \pi_i(t)]^2\cdot I\left(Y_{i1} > t,\ Y_{i2} > t\right\}}{G_i(t)}.  
\end{split}
\end{align}

\vspace{1ex}

\noindent If $G_i(t)$ is known, the expectation of the IPCW-approximated bivariate Brier score is equal to ${\rm MSE}(t)$ plus a constant that is free of $\pi_i(t)$, which represents the irreducible error incurred by approximating $S_i(t)$ in a  data-driven fashion (Supplement C). As $G_i(t)$ is unknown in practice,  we replace it  by $\hat{G}_i(t)$, its  Kaplan-Meier estimate.

\subsection{\textcolor{black}{Bivariate Concordance Index}}

\textcolor{black}{We further propose a bivariate extension to Harrell's concordance (C) index to evaluate the discriminative ability of the predicted hazards in a semi-competing risks setting \citep{harrell1996multivariable}. For each individual, we transformed the predicted hazard values for each transition into standardized scores by computing their empirical cumulative distribution function (ECDF) ranks and then applying the inverse normal transformation. The resulting quantile scores were averaged to obtain an overall risk score for each subject. We then considered all pairs of individuals $(i, j)$ such that at least one of them experienced a non-censoring event (i.e., $D_{1i} = 1$ or $D_{2i} = 1$, and similarly for subject $j$). A pair was defined as comparable if the event time of subject $i$ was earlier than that of subject $j$ in either the non-terminal or terminal event ($Y_{1i} < Y_{1j}$ or $Y_{2i} < Y_{2j}$). A comparable pair was considered concordant if the subject with the earlier event time also had a higher predicted risk score. The bivariate concordance index was then computed as the proportion of concordant pairs among all comparable pairs.}

%--- SECTION 5 ------------------------------------------------------------------

\section{\textcolor{black}{Simulation Studies}}
\label{sec:5}

\noindent \textcolor{black}{We conducted a series of simulations to validate our neural EM algorithm and illustrate the feasibility of our method against existing approaches. We simulated data from Equation (\ref{eq:condlik}), varying sample size, the population frailty variance, log-risk functions, and censoring rates across 16 settings. In particular, we generated independent datasets with sample sizes of either $n = 1,000$ or $n = 10,000$. We then simulated the $i$th individual's frailty, $\gamma_i$, from a Gamma distribution with mean 1 and variance $\theta$, taking $\theta$ to be 0.5 or 2.0, corresponding to varying degrees of dependence between event times. In a separate sensitivity analysis, we further explored the Gamma frailty assumption (see section \ref{sec:gamma_sensitivity} below). To generate the risk functions, $h_g(\boldsymbol{X}_i); g \in \{1,2,3\}$, we considered linear and nonlinear functions of our simulated risk factors. In both cases, we generated twelve covariates, as in our motivating data, from a multivariate Normal distribution with a zero mean vector and a compound symmetric covariance matrix with diagonal elements (variances) equal to one and off-diagonal elements (covariances) equal to 0.2. That is,}

\textcolor{black}{
\[
\boldsymbol{X}_i \sim \mathcal{N}_{12}\left(
  \begin{bmatrix} 0 \\ 0 \\ \vdots \\ 0
  \end{bmatrix},\ 
  \begin{bmatrix}
    1.0    & 0.2    & \cdots & 0.2    \\
    0.2    & 1.0    & \cdots & 0.2    \\
    \vdots & \vdots & \ddots & \vdots \\
    0.2    & 0.2    & \dots  & 1.0
  \end{bmatrix}
\right).
\]
}

\subsection{\textcolor{black}{Linear Log-Risk Functions}}

\textcolor{black}{In the first simulation scenario, we considered a linear form for the log-risk functions so the requirements for the classical models were satisfied, facilitating a fair comparison. We generated $h_g(\boldsymbol{X}_i), g\in \{1,2,3\}$ so that }

\textcolor{black}{
\[
        h_g(\boldsymbol{X}_i) = X_{i,1}\beta_{1,g} + \cdots + X_{i,12}\beta_{12,g},
        \]
 with $(\beta_{1,g}, \ldots, \beta_{12, g}) = (-0.5, -0.5, -0.5, -0.5, 0.5, 0.5, 0.5, 0.5, 0.5, 0, 0, 0),$
}
\noindent \textcolor{black}{
that is, we assumed that the first nine covariates $(X_1, \ldots, X_9)$ were related to each risk function. 
%We fixed $\boldsymbol{\beta}_g$ across all simulation replicates. 
We further generated the censoring times, $C_i$, to be covariate-dependent, coming from an exponential distribution with hazard}

\textcolor{black}{
\[        
    \lambda_c(\boldsymbol{X}_i) = \mu_C \times \exp\{X_{i,1}\alpha_1 + \cdots + X_{i,12}\alpha_{12}\}, 
\]
where $(\alpha_{1}, \ldots, \alpha_{12}) = (0, 0, 0, 0, 0, 0, 0, -0.5, 0.5, -0.5, 0.5, -0.5)$, so that six covariates were related to the censoring time, with three related to the survival and censoring times. Lastly, we selected $\mu_C$ to achieve approximate censoring rates of 25\% or 50\%.}

\subsection{\textcolor{black}{Non-Linear Log-Risk Functions}}

\textcolor{black}{In the second simulation scenario, we generated nonlinear risk functions to highlight the utility of our method. Namely, we generated}

\textcolor{black}{
\[
\begin{aligned}
  h_g(\boldsymbol{X}_i) &= \beta_{1,g} \exp(X_{i,1} - X_{i,2}) - \beta_{2,g} \log\{(X_{i,3} + X_{i,4})^2 \} \\ &\quad +\ \beta_{3,g} \sin(X_{i,5} X_{i,6}) - \beta_{4,g} (X_{i,7} - X_{i,8} + X_{i,9})^2,\\
  \boldsymbol{\beta}_g &= (\beta_{1,g}, \ldots, \beta_{4,g}) = (=-0.5, -0.5, 0.5, 0.5).
\end{aligned}
\]
}

\noindent \textcolor{black}{we again generated censoring times, $C_i$, as before, where $\mu_C$ was chosen to target approximate censoring rates of 25\% or 50\%. Across all simulation settings, we generated baseline hazard functions, $\lambda_{01}$, $\lambda_{02}$, and $\lambda_{03}$, from exponential distributions so that $\lambda_{01}$ = $\lambda_{03}$ = 2, and $\lambda_{02}$ = 3. For each parameter configuration, we generated 500 independent datasets.} 

\textcolor{black}{We compared our method to five existing methods for analyzing semi-competing risks, namely those proposed by \cite{xu2010statistical, lee2015bayesian, lee2017accelerated, gorfine2021marginalized}, and \cite{kats2022accelerated}. We followed the authors' original implementations and adopted recommended default settings when available. In choosing the prior hyperparameters for the two Bayesian methods under comparison, we opted for relatively flat priors for the six Weibull baseline hazard hyperparameters, corresponding to the three parametric baseline hazards. These initial hyperparameters are recommended as the defaults settings in the implementation of these methods. For the prior mean and standard deviation of the frailty variance, we tuned possible choices over a factorial grid for every combination of these hyperparameters ranging from 0.1 to 1.0 in increments of 0.1 (100 combinations). We  chose the configuration which yielded the best predictive performance and carried this forward. This ensured a fair comparison by allowing each method to perform under its optimal or near-optimal configuration, given the same input data structure.}

\textcolor{black}{For our method, we varied the number of nodes per layer, the dropout fraction, the degree of regularization, and learning rate over a grid of values to determine the setting with the best predictive performance. Specifically, we choose three hidden layers, we selected the nodes per layer from 16 to 1024, we set the dropout fraction to be 0.3, and we tuned the learning rate from 0.0001 to 0.05. We assessed performance via the mean (SD) estimated frailty variance parameter, the mean (SD) bivariate Brier score integrated up to $t = 1$ year, the mean (SD) bivariate concordance index, the mean (SD) bias for estimating the baseline hazard functions at $t = 1$, and the average (SD) mean integrated squared error (MISE) for estimating the log-risk surfaces for each state transition hazard, separately:}
\[
    {\rm MISE}_g = \frac{1}{n} \sum_{i=1}^n \left[h_g(\boldsymbol{X}_i) - \hat{h}_g(\boldsymbol{X}_i)\right]^2;\ g = 1,2,3.
\]

\subsection{\textcolor{black}{Results}}

\textcolor{black}{Tables \ref{tab:sim_theta}-\ref{tab:sim_cindex} summarize the results of this simulation study. Table \ref{tab:sim_theta} gives the mean (SD) results for estimating the frailty variance, $\theta$, averaged over 500 independent replicates in each simulation setting. As shown, the proposed method has the lowest bias in 12 of the 16 settings, specifically in the eight settings where the true risk functions are non-linear and in four of the eight linear settings. Among the four linear settings where the proposed method does not have the lowest bias, the results are comparable to the methods of \cite{lee2017accelerated} and \cite{gorfine2021marginalized}, which have the best performances. We also note that the estimated $\hat{\theta}$ are slightly closer to the truth for smaller values of $\theta$. Table \ref{tab:sim_mise} then compares each approach in terms of the MISE for the predicted log-risk functions. Across all methods, the MISE increases slightly with the frailty variance and censoring rate. In addition, the variability decreases with increasing sample size. Despite being a highly nonlinear approximation, the proposed method performs comparably to the better performing methods, across all state transitions, when the true underlying function of the predictors is linear. In all nonlinear settings, our approach has the lowest MISEs, suggesting that our method outperforms the (semi-)parametric approaches when the functional form of the predictors is truly nonlinear. We also study the estimation of the baseline hazard functions across all state transitions in Table \ref{tab:sim_baseline}. These results show that the proposed method has the lowest bias in 10 of the 16 cases, including all eight with nonlinear covariate effects. Moreover, the proposed method has a bias comparable to the better performing methods in settings where it does not have the lowest bias, with more variation shown as the censoring rate increases.} 

\textcolor{black}{Lastly, to assess the predictive accuracy of the proposed method, we study both the integrated bivariate Brier score and the bivariate concordance (C) index, as described above. We calculate the integrated bivariate Brier score for one year survival over a sequence of 100 evenly spaced time points in each simulation and compared the results of our method with the existing methods. As shown, the proposed method has the lowest integrated bivariate Brier score (lower values indicating higher predictive accuracy) in 13 of the 16 settings (Table \ref{tab:sim_bbs}). Similarly, the proposed method has the highest bivariate C-index (higher values indicating higher predictive accuracy) in 14 of 16 simulation settings, and a comparable bivariate C-index in the two settings where it is not strictly the highest (Table \ref{tab:sim_cindex}).} 

\begin{table}[!ht]
\tiny
\caption{\textcolor{black}{Estimated mean frailty variance and empirical standard errors (in parentheses), averaged over 500 replicates for each simulation setting. Bold values denote the method which has an average value closest to the true value of $\theta$.}}
\label{tab:sim_theta}
\centering
\vspace{2ex}
{\color{black}
\groupedRowColors[0]{4}{3}{white}{gray!10}
\begin{tabular}{cccccccccc}
\toprule
\multicolumn{4}{c}{Simulation Settings} & \multicolumn{6}{c}{Methods} \\
\cmidrule(lr){1-4} \cmidrule(lr){5-10}
\rowcolor{white}
Sample Size & Risk Function & Censoring Rate & $\theta$ & Xu (2010) & Lee (2015) & Lee (2017) & Gorfine (2020) & Kats (2022) & Proposed \\
\midrule
1,000 & Linear     & 25\% & 0.5 & 0.43 (0.09) & 0.93 (0.05) & 0.54 (0.26) &      0.76 (0.26)  & 0.65 (0.12) & {\bf 0.51 (0.03)} \\
5,000 & Linear     & 25\% & 0.5 & 0.42 (0.04) & 0.92 (0.04) & 0.55 (0.26) &      0.48 (0.11)  & 0.68 (0.12) & {\bf 0.50 (0.02)} \\
1,000 & Linear     & 25\% & 2.0 & 1.45 (0.18) & 0.79 (0.06) & 0.57 (0.26) & {\bf 2.04 (0.57)} & 1.40 (0.19) &      2.06 (0.03)  \\
5,000 & Linear     & 25\% & 2.0 & 1.44 (0.08) & 0.78 (0.06) & 0.57 (0.25) & {\bf 1.97 (0.21)} & 1.48 (0.08) &      2.08 (0.01)  \\
1,000 & Non-Linear & 25\% & 0.5 & 0.70 (0.05) & 0.52 (0.17) & 0.57 (0.26) &      1.74 (0.39)  & 2.59 (3.12) & {\bf 0.49 (0.01)} \\
5,000 & Non-Linear & 25\% & 0.5 & 0.71 (0.02) & 0.70 (0.05) & 0.55 (0.25) &      1.48 (0.17)  & 2.21 (1.67) & {\bf 0.46 (0.01)} \\
1,000 & Non-Linear & 25\% & 2.0 & 0.83 (0.26) & 0.73 (0.28) & 0.55 (0.27) &      3.12 (0.62)  & 4.93 (5.79) & {\bf 2.14 (0.11)} \\
5,000 & Non-Linear & 25\% & 2.0 & 0.81 (0.11) & 0.73 (0.26) & 0.56 (0.26) &      2.83 (0.26)  & 4.52 (3.96) & {\bf 2.11 (0.01)} \\
\midrule
1,000 & Linear     & 50\% & 0.5 & 0.37 (0.10) & 0.42 (0.13) &      0.58 (0.27)  &      1.12 (0.41)  & 1.25 (0.21) & {\bf 0.57 (0.02)} \\
5,000 & Linear     & 50\% & 0.5 & 0.36 (0.04) & 0.40 (0.12) & {\bf 0.55 (0.26)} &      0.56 (0.15)  & 1.31 (0.10) &      0.58 (0.01)  \\
1,000 & Linear     & 50\% & 2.0 & 1.15 (0.18) & 0.65 (0.23) &      0.58 (0.26)  &      2.88 (0.83)  & 2.54 (0.51) & {\bf 2.13 (0.02)} \\
5,000 & Linear     & 50\% & 2.0 & 1.13 (0.08) & 0.64 (0.22) &      0.56 (0.26)  & {\bf 2.01 (0.29)} & 2.68 (0.16) &      2.10 (0.01)  \\ 
1,000 & Non-Linear & 50\% & 0.5 & 1.76 (1.00) & 0.54 (0.20) &      0.60 (0.27)  &      2.53 (0.76)  & 3.65 (2.50) & {\bf 0.52 (0.01)} \\
5,000 & Non-Linear & 50\% & 0.5 & 0.34 (0.06) & 0.57 (0.19) &      0.57 (0.26)  &      1.62 (0.32)  & 3.37 (1.99) & {\bf 0.54 (0.01)} \\
1,000 & Non-Linear & 50\% & 2.0 & 0.99 (0.39) & 0.67 (0.26) &      0.56 (0.26)  &      3.59 (0.85)  & 4.94 (4.29) & {\bf 2.09 (0.15)} \\
5,000 & Non-Linear & 50\% & 2.0 & 0.59 (0.10) & 0.68 (0.26) &      0.59 (0.27)  &      2.95 (0.34)  & 4.14 (2.58) & {\bf 2.03 (0.01)} \\
\bottomrule
\end{tabular}
}
\end{table}

\begin{table}[!ht]
\tiny
\caption{\textcolor{black}{Average (SD) mean integrated squared errors (MISE) for the simulated log-risk surfaces, $h_g(\boldsymbol{X}_i)$, for each state transition hazard; $g = 1, 2, 3$, averaged over 500 replicates for each simulation setting. Bold values denote the method which has the lowest average MISE for each setting and state transition.}}
\label{tab:sim_mise}
\centering
\vspace{2ex}
{\color{black}
\groupedRowColors[0]{4}{4}{white}{gray!10}
\begin{tabular}{cccccccccc}
\toprule
\multicolumn{4}{c}{Simulation Settings} & \multicolumn{6}{c}{Methods} \\
\cmidrule(lr){1-4} \cmidrule(lr){5-10}
Sample Size & Risk Function & Censoring Rate & $\theta$ & Xu (2010) & Lee (2015) & Lee (2017) & Gorfine (2020) & Kats (2022) & Proposed \\
\midrule
\multicolumn{10}{c}{\bf First Transition: $h_1(\boldsymbol{X}_i)$} \\
\midrule
1,000 & Linear     & 25\% & 0.5 & 0.15 (0.05)  & 0.41 (0.09) &      0.26 (0.09)  & {\bf 0.13 (0.05)} & 2.97 (0.29) &      0.36 (0.11)  \\
5,000 & Linear     & 25\% & 0.5 & 0.12 (0.02)  & 0.32 (0.04) &      0.18 (0.04)  & {\bf 0.11 (0.02)} & 5.70 (0.33) &      0.28 (0.06)  \\
1,000 & Linear     & 25\% & 2.0 & 0.10 (0.04)  & 0.58 (0.10) & {\bf 0.26 (0.09)} &      0.56 (0.09)  & 5.70 (0.65) &      0.33 (0.11)  \\
5,000 & Linear     & 25\% & 2.0 & 0.04 (0.02)  & 0.47 (0.06) & {\bf 0.16 (0.03)} &      0.54 (0.04)  & 5.38 (0.49) &      0.22 (0.04)  \\
1,000 & Non-Linear & 25\% & 0.5 & 4.27 (0.39)  & 4.47 (0.40) &      4.37 (0.38)  &      4.49 (0.41)  & 8.04 (0.98) & {\bf 4.01 (0.58)} \\
5,000 & Non-Linear & 25\% & 0.5 & 4.26 (0.17)  & 4.41 (0.18) &      4.27 (0.17)  &      4.48 (0.18)  & 8.01 (0.68) & {\bf 2.94 (0.23)} \\
1,000 & Non-Linear & 25\% & 2.0 & 4.31 (0.41)  & 4.54 (0.41) &      4.43 (0.40)  &      4.56 (0.41)  & 7.99 (1.14) & {\bf 3.82 (0.59)} \\
5,000 & Non-Linear & 25\% & 2.0 & 4.31 (0.18)  & 4.49 (0.18) &      4.33 (0.17)  &      4.58 (0.18)  & 8.06 (0.66) & {\bf 2.88 (0.20)} \\
\midrule
1,000 & Linear     & 50\% & 0.5 & {\bf 0.46 (0.10)} & 0.70 (0.11) & 0.58 (0.15) &      2.47 (0.26)  & 5.63 (0.63) &      0.73 (0.17)  \\ 
5,000 & Linear     & 50\% & 0.5 &      0.43 (0.04)  & 0.62 (0.05) & 0.50 (0.07) & {\bf 0.09 (0.02)} & 5.40 (0.24) &      0.78 (0.08)  \\ 
1,000 & Linear     & 50\% & 2.0 &      0.46 (0.12)  & 0.89 (0.12) & 0.71 (0.16) & {\bf 0.45 (0.10)} & 5.59 (0.76) &      0.88 (0.16)  \\ 
5,000 & Linear     & 50\% & 2.0 & {\bf 0.41 (0.06)} & 0.80 (0.08) & 0.63 (0.07) &      0.42 (0.08)  & 5.09 (0.31) &      0.79 (0.07)  \\ 
1,000 & Non-Linear & 50\% & 0.5 &      5.14 (0.68)  & 4.66 (0.42) & 4.78 (0.43) &      4.40 (0.40)  & 8.10 (1.02) & {\bf 4.07 (0.52)} \\
5,000 & Non-Linear & 50\% & 0.5 &      4.58 (0.19)  & 4.60 (0.19) & 4.68 (0.19) &      4.36 (0.17)  & 8.06 (0.67) & {\bf 3.57 (0.18)} \\
1,000 & Non-Linear & 50\% & 2.0 &      4.94 (0.63)  & 4.64 (0.42) & 4.66 (0.42) &      4.50 (0.40)  & 8.12 (1.04) & {\bf 4.15 (0.48)} \\
5,000 & Non-Linear & 50\% & 2.0 &      4.50 (0.19)  & 4.59 (0.19) & 4.58 (0.18) &      4.51 (0.18)  & 8.10 (0.67) & {\bf 3.31 (0.19)} \\
\midrule
\multicolumn{10}{c}{\bf Second Transition: $h_2(\boldsymbol{X}_i)$} \\
\midrule
1,000 & Linear     & 25\% & 0.5 & {\bf 0.04 (0.02)}  & 0.17 (0.05) & 0.08 (0.04) & 0.12 (0.04) & 2.92 (0.23) &      0.17 (0.08)  \\
5,000 & Linear     & 25\% & 0.5 & {\bf 0.01 (0.004)} & 0.12 (0.02) & 0.02 (0.01) & 0.11 (0.02) & 5.69 (0.31) &      0.09 (0.03)  \\
1,000 & Linear     & 25\% & 2.0 & {\bf 0.08 (0.03)}  & 0.45 (0.08) & 0.13 (0.05) & 0.54 (0.08) & 5.59 (0.50) &      0.29 (0.11)  \\
5,000 & Linear     & 25\% & 2.0 & {\bf 0.03 (0.01)}  & 0.36 (0.05) & 0.05 (0.02) & 0.54 (0.03) & 5.36 (0.46) &      0.17 (0.04)  \\
1,000 & Non-Linear & 25\% & 0.5 &      4.24 (0.38)   & 4.42 (0.38) & 4.21 (0.36) & 4.47 (0.39) & 7.92 (0.93) & {\bf 4.04 (0.60)} \\
5,000 & Non-Linear & 25\% & 0.5 &      4.24 (0.17)   & 4.38 (0.17) & 4.16 (0.16) & 4.48 (0.17) & 7.93 (0.45) & {\bf 3.16 (0.27)} \\
1,000 & Non-Linear & 25\% & 2.0 &      4.28 (0.40)   & 4.47 (0.40) & 4.25 (0.37) & 4.55 (0.40) & 7.75 (0.98) & {\bf 3.77 (0.57)} \\
5,000 & Non-Linear & 25\% & 2.0 &      4.29 (0.17)   & 4.45 (0.18) & 4.20 (0.16) & 4.58 (0.18) & 7.78 (0.45) & {\bf 2.95 (0.23)} \\
\midrule
1,000 & Linear     & 50\% & 0.5 & {\bf 0.05 (0.02)} & 0.16 (0.05) &      0.10 (0.05)  & 2.45 (0.25) & 5.53 (0.46) &      0.26 (0.14)  \\ 
5,000 & Linear     & 50\% & 0.5 & {\bf 0.02 (0.01)} & 0.10 (0.02) &      0.04 (0.02)  & 0.08 (0.02) & 5.39 (0.19) &      0.16 (0.05)  \\
1,000 & Linear     & 50\% & 2.0 & {\bf 0.12 (0.05)} & 0.39 (0.08) &      0.16 (0.06)  & 0.43 (0.08) & 5.39 (0.54) &      0.35 (0.11)  \\
5,000 & Linear     & 50\% & 2.0 & {\bf 0.07 (0.02)} & 0.31 (0.04) & {\bf 0.07 (0.03)} & 0.41 (0.07) & 5.02 (0.23) &      0.21 (0.04)  \\
1,000 & Non-Linear & 50\% & 0.5 &      5.01 (0.81)  & 4.35 (0.37) &      4.31 (0.37)  & 4.37 (0.38) & 7.93 (0.94) & {\bf 4.28 (0.55)} \\
5,000 & Non-Linear & 50\% & 0.5 &      4.27 (0.17)  & 4.31 (0.17) &      4.21 (0.17)  & 4.36 (0.17) & 7.92 (0.45) & {\bf 3.42 (0.32)} \\
1,000 & Non-Linear & 50\% & 2.0 &      4.84 (0.70)  & 4.44 (0.39) &      4.32 (0.38)  & 4.49 (0.39) & 7.90 (0.96) & {\bf 4.03 (0.58)} \\
5,000 & Non-Linear & 50\% & 2.0 &      4.33 (0.17)  & 4.41 (0.17) &      4.26 (0.17)  & 4.51 (0.18) & 7.91 (0.46) & {\bf 3.16 (0.25)} \\
\midrule
\multicolumn{10}{c}{\bf Third Transition: $h_3(\boldsymbol{X}_i)$} \\
\midrule
1,000 & Linear     & 25\% & 0.5 & {\bf 0.15 (0.05)} & 0.41 (0.09) &      0.16 (0.07)  & 0.13 (0.05) & 2.97 (0.29) &      0.33 (0.12)  \\
5,000 & Linear     & 25\% & 0.5 &      0.12 (0.02)  & 0.32 (0.04) & {\bf 0.05 (0.02)} & 0.11 (0.02) & 5.70 (0.33) &      0.15 (0.05)  \\
1,000 & Linear     & 25\% & 2.0 & {\bf 0.10 (0.04)} & 0.58 (0.10) &      0.29 (0.12)  & 0.56 (0.09) & 5.70 (0.65) &      0.40 (0.14)  \\
5,000 & Linear     & 25\% & 2.0 & {\bf 0.04 (0.02)} & 0.47 (0.06) &      0.13 (0.04)  & 0.54 (0.04) & 5.38 (0.50) &      0.25 (0.06)  \\
1,000 & Non-Linear & 25\% & 0.5 &      4.27 (0.39)  & 4.47 (0.40) &      4.52 (0.43)  & 4.49 (0.41) & 8.04 (0.98) & {\bf 4.19 (0.49)} \\
5,000 & Non-Linear & 25\% & 0.5 &      4.26 (0.17)  & 4.41 (0.18) &      4.35 (0.18)  & 4.48 (0.18) & 8.01 (0.68) & {\bf 3.27 (0.25)} \\
1,000 & Non-Linear & 25\% & 2.0 &      4.31 (0.41)  & 4.54 (0.41) &      4.64 (0.45)  & 4.56 (0.41) & 7.99 (1.14) & {\bf 4.05 (0.56)} \\
5,000 & Non-Linear & 25\% & 2.0 &      4.31 (0.18)  & 4.49 (0.18) &      4.45 (0.19)  & 4.58 (0.18) & 8.06 (0.66) & {\bf 3.17 (0.24)} \\
\midrule
1,000 & Linear     & 50\% & 0.5 & 0.46 (0.10)  & 0.70 (0.11) & {\bf 0.23 (0.10)} & 2.47 (0.26) & 5.63 (0.63) &      0.43 (0.18)  \\ 
5,000 & Linear     & 50\% & 0.5 & 0.43 (0.05)  & 0.61 (0.06) & {\bf 0.08 (0.03)} & 0.09 (0.02) & 5.40 (0.24) &      0.31 (0.12)  \\ 
1,000 & Linear     & 50\% & 2.0 & 0.46 (0.12)  & 0.89 (0.12) & {\bf 0.32 (0.13)} & 0.45 (0.10) & 5.59 (0.76) &      0.61 (0.18)  \\ 
5,000 & Linear     & 50\% & 2.0 & 0.41 (0.06)  & 0.80 (0.08) & {\bf 0.14 (0.05)} & 0.42 (0.08) & 5.09 (0.32) &      0.51 (0.09)  \\ 
1,000 & Non-Linear & 50\% & 0.5 & 5.14 (0.68)  & 4.66 (0.42) &      4.70 (0.45)  & 4.40 (0.40) & 8.10 (1.02) & {\bf 4.39 (0.58)} \\
5,000 & Non-Linear & 50\% & 0.5 & 4.57 (0.19)  & 4.60 (0.19) &      4.40 (0.19)  & 4.36 (0.17) & 8.06 (0.67) & {\bf 3.42 (0.27)} \\
1,000 & Non-Linear & 50\% & 2.0 & 4.94 (0.63)  & 4.64 (0.42) &      4.68 (0.46)  & 4.50 (0.40) & 8.12 (1.05) & {\bf 4.47 (0.54)} \\
5,000 & Non-Linear & 50\% & 2.0 & 4.50 (0.19)  & 4.59 (0.19) &      4.45 (0.20)  & 4.51 (0.18) & 8.10 (0.67) & {\bf 3.56 (0.28)} \\
\bottomrule
\end{tabular}
}
\end{table}

\begin{table}[!ht]
\tiny
\caption{\textcolor{black}{Average bias (SD) for estimating the baseline hazards, $\lambda_{0g}(t); g = 1,2,3$, evaluated at $t = 1$ and averaged over all $g$ baselines. Bold values denote the method which has the lowest average bias for each simulation setting.}}
\label{tab:sim_baseline}
\centering
\vspace{2ex}
{\color{black}
\groupedRowColors[0]{4}{3}{white}{gray!10}
\begin{tabular}{cccccccccc}
\toprule
\multicolumn{4}{c}{Simulation Settings} & \multicolumn{6}{c}{Methods} \\
\cmidrule(lr){1-4} \cmidrule(lr){5-10}
\rowcolor{white}
Sample Size & Risk Function & Censoring Rate & $\theta$ & \textcolor{black}{Xu (2010)} & \textcolor{black}{Lee (2015)} & \textcolor{black}{Lee (2017)} & \textcolor{black}{Gorfine (2020)} & \textcolor{black}{Kats (2022)} & Proposed \\
\midrule
1,000 & Linear     & 25\% & 0.5 &      0.28 (0.50)  & {\bf 0.20 (0.30)} & 10.46 (3.52) & 0.96 (0.19) & 4.36 (2.71) &      0.27 (0.44)  \\
5,000 & Linear     & 25\% & 0.5 & {\bf 0.15 (0.21)} &      0.19 (0.29)  & 10.52 (3.61) & 1.05 (0.89) & 4.42 (3.12) &      0.17 (0.25)  \\
1,000 & Linear     & 25\% & 2.0 & {\bf 0.36 (0.53)} &      1.62 (1.59)  &  7.37 (3.50) & 2.37 (0.10) & 4.20 (2.45) &      0.57 (0.58)  \\
5,000 & Linear     & 25\% & 2.0 & {\bf 0.32 (0.44)} &      1.62 (1.59)  &  7.39 (3.49) & 2.62 (2.09) & 4.24 (3.17) &      0.47 (0.39)  \\
1,000 & Non-Linear & 25\% & 0.5 &      1.00 (0.91)  &      1.08 (0.94)  & 11.14 (4.16) & 1.56 (0.12) & 4.62 (4.86) & {\bf 0.25 (0.47)} \\
5,000 & Non-Linear & 25\% & 0.5 &      0.92 (0.72)  &      1.07 (0.87)  & 11.11 (4.20) & 2.41 (0.84) & 4.38 (2.37) & {\bf 0.37 (0.31)} \\
1,000 & Non-Linear & 25\% & 2.0 &      1.72 (1.31)  &      1.94 (1.41)  &  8.79 (4.07) & 2.37 (0.09) & 4.21 (2.46) & {\bf 0.22 (0.46)} \\
5,000 & Non-Linear & 25\% & 2.0 &      1.70 (1.20)  &      1.91 (1.36)  &  8.82 (4.03) & 3.30 (1.50) & 4.25 (2.44) & {\bf 0.12 (0.08)} \\
\midrule
1,000 & Linear     & 50\% & 0.5 &      1.40 (2.38)  & {\bf 0.94 (1.39)} & 10.83 (3.28) & 1.17 (0.98) & 4.69 (2.32) &      0.95 (1.49)  \\
5,000 & Linear     & 50\% & 0.5 &      1.00 (1.44)  &      0.84 (1.19)  & 10.92 (3.32) & 1.07 (0.90) & 4.64 (2.30) & {\bf 0.46 (0.57)} \\
1,000 & Linear     & 50\% & 2.0 & {\bf 1.02 (1.31)} &      1.15 (1.55)  &  9.02 (2.88) & 2.82 (2.00) & 4.58 (2.23) &      1.12 (2.48)  \\
5,000 & Linear     & 50\% & 2.0 &      0.78 (0.68)  &      1.12 (1.54)  &  9.11 (2.94) & 2.65 (2.14) & 4.60 (2.38) & {\bf 0.57 (0.69)} \\
1,000 & Non-Linear & 50\% & 0.5 &      2.39 (1.54)  &      1.45 (1.17)  & 12.14 (3.32) & 3.31 (2.37) & 4.66 (2.36) & {\bf 0.31 (0.27)} \\
5,000 & Non-Linear & 50\% & 0.5 &      1.15 (0.71)  &      1.39 (0.97)  & 12.15 (3.17) & 2.44 (0.85) & 4.73 (2.27) & {\bf 0.17 (0.20)} \\
1,000 & Non-Linear & 50\% & 2.0 &      2.51 (1.58)  &      1.74 (1.45)  & 10.00 (3.66) & 3.71 (1.64) & 4.46 (2.31) & {\bf 0.27 (0.45)} \\
5,000 & Non-Linear & 50\% & 2.0 &      1.79 (1.40)  &      1.70 (1.38)  & 10.08 (3.66) & 3.34 (1.47) & 4.57 (2.42) & {\bf 0.13 (0.15)} \\
\bottomrule
\end{tabular}
}
\end{table}

\begin{table}[!ht]
\tiny
\caption{\textcolor{black}{Average (SD) one year integrated bivarate Brier scores (iBBS). Bold values denote the method which has the lowest (lower values = higher predictive accuracy) average iBBS for each simulation setting.}}
\label{tab:sim_bbs}
\centering
\vspace{2ex}
{\color{black}
\groupedRowColors[0]{4}{3}{white}{gray!10}
\begin{tabular}{cccccccccc}
\toprule
\multicolumn{4}{c}{Simulation Settings} & \multicolumn{6}{c}{Methods} \\
\cmidrule(lr){1-4} \cmidrule(lr){5-10}
\rowcolor{white}
Sample Size & Risk Function & Censoring Rate & $\theta$ & Xu (2010) & Lee (2015) & Lee (2017) & Gorfine (2020) & Kats (2022) & Proposed \\
\midrule
1,000 & Linear     & 25\% & 0.5 &      0.09 (0.005)  & 0.09 (0.006) & 0.10 (0.006) & 0.07 (0.01) & 0.47 (0.02) & {\bf 0.05 (0.003)} \\
5,000 & Linear     & 25\% & 0.5 &      0.09 (0.002)  & 0.09 (0.004) & 0.09 (0.005) & 0.06 (0.004)& 0.50 (0.02) & {\bf 0.05 (0.001)} \\
1,000 & Linear     & 25\% & 2.0 &      0.23 (0.009)  & 0.21 (0.010) & 0.23 (0.010) & 0.21 (0.05) & 0.51 (0.02) & {\bf 0.16 (0.006)} \\
5,000 & Linear     & 25\% & 2.0 & {\bf 0.16 (0.003)} & 0.22 (0.007) & 0.20 (0.006) & 0.26 (0.01) & 0.51 (0.02) & {\bf 0.16 (0.003)} \\
1,000 & Non-Linear & 25\% & 0.5 &      0.15 (0,009)  & 0.14 (0.009) & 0.14 (0.010) & 0.18 (0.03) & 0.52 (0.13) & {\bf 0.07 (0.005)} \\
5,000 & Non-Linear & 25\% & 0.5 &      0.15 (0.004)  & 0.13 (0.007) & 0.16 (0.005) & 0.17 (0.01) & 0.51 (0.10) & {\bf 0.06 (0.003)} \\
1,000 & Non-Linear & 25\% & 2.0 &      0.24 (0.010)  & 0.18 (0.007) & 0.21 (0.008) & 0.33 (0.03) & 0.55 (0.11) & {\bf 0.16 (0.006)} \\
5,000 & Non-Linear & 25\% & 2.0 &      0.18 (0.003)  & 0.18 (0.003) & 0.19 (0.003) & 0.32 (0.02) & 0.52 (0.09) & {\bf 0.10 (0.005)} \\
\midrule
1,000 & Linear     & 50\% & 0.5 &      0.06 (0.009)  & {\bf 0.03 (0.005)} &      0.05 (0.006)  & 0.11 (0.02) & 0.58 (0.03) & {\bf 0.03 (0.005)} \\
5,000 & Linear     & 50\% & 0.5 & {\bf 0.03 (0.001)} & {\bf 0.03 (0.001)} & {\bf 0.03 (0.001)} & 0.06 (0.01) & 0.58 (0.02) & {\bf 0.03 (0.002)} \\
1,000 & Linear     & 50\% & 2.0 & {\bf 0.03 (0.001)} &      0.10 (0.007)  &      0.07 (0.003)  & 0.31 (0.05) & 0.61 (0.03) &      0.04 (0.006)  \\ 
5,000 & Linear     & 50\% & 2.0 &      0.10 (0.003)  &      0.10 (0.003)  &      0.10 (0.003)  & 0.27 (0.02) & 0.61 (0.02) & {\bf 0.02 (0.001)} \\ 
1,000 & Non-Linear & 50\% & 0.5 &      0.15 (0.067)  & {\bf 0.05 (0.006)} &      0.08 (0.009)  & 0.22 (0.05) & 0.62 (0.09) &      0.14 (0.004)  \\
5,000 & Non-Linear & 50\% & 0.5 & {\bf 0.05 (0.003)} & {\bf 0.05 (0.003)} & {\bf 0.05 (0.003)} & 0.17 (0.02) & 0.54 (0.06) &      0.09 (0.004)  \\
1,000 & Non-Linear & 50\% & 2.0 &      0.19 (0.016)  & {\bf 0.13 (0.007)} &      0.17 (0.011)  & 0.36 (0.04) & 0.61 (0.10) & {\bf 0.13 (0.010)} \\
5,000 & Non-Linear & 50\% & 2.0 &      0.13 (0.003)  &      0.13 (0.004)  &      0.13 (0.004)  & 0.33 (0.03) & 0.58 (0.07) & {\bf 0.09 (0.005)} \\
\bottomrule
\end{tabular}
}
\end{table}

\begin{table}[!ht]
\tiny
\caption{\textcolor{black}{Average (SD) bivarate concordance (C) index. Bold values denote the method which has the highest (higher values = higher predictive accuracy) average bivarate C-index for each simulation setting.}}
\label{tab:sim_cindex}
\centering
\vspace{2ex}
{\color{black}
\groupedRowColors[0]{4}{3}{white}{gray!10}
\begin{tabular}{cccccccccc}
\toprule
\multicolumn{4}{c}{Simulation Settings} & \multicolumn{6}{c}{Methods} \\
\cmidrule(lr){1-4} \cmidrule(lr){5-10}
\rowcolor{white}
Sample Size & Risk Function & Censoring Rate & $\theta$ & Xu (2010) & Lee (2015) & Lee (2017) & Gorfine (2020) & Kats (2022) & Proposed \\
\midrule
1,000 & Linear     & 25\% & 0.5 & {\bf 0.67 (0.01)} &      0.66 (0.01)  & 0.53 (0.02) & 0.64 (0.02) & 0.50 (0.02) & {\bf 0.67 (0.01)} \\
5,000 & Linear     & 25\% & 0.5 &      0.63 (0.01)  & {\bf 0.68 (0.01)} & 0.56 (0.02) & 0.66 (0.01) & 0.34 (0.02) &      0.67 (0.01)  \\
1,000 & Linear     & 25\% & 2.0 &      0.63 (0.01)  &      0.63 (0.02)  & 0.51 (0.02) & 0.61 (0.06) & 0.41 (0.02) & {\bf 0.64 (0.01)} \\
5,000 & Linear     & 25\% & 2.0 &      0.63 (0.01)  &      0.62 (0.01)  & 0.51 (0.02) & 0.54 (0.01) & 0.42 (0.02) & {\bf 0.64 (0.01)} \\
1,000 & Non-Linear & 25\% & 0.5 &      0.61 (0.01)  &      0.61 (0.01)  & 0.52 (0.02) & 0.53 (0.02) & 0.43 (0.03) & {\bf 0.72 (0.03)} \\
5,000 & Non-Linear & 25\% & 0.5 &      0.64 (0.01)  &      0.62 (0.02)  & 0.54 (0.01) & 0.53 (0.01) & 0.45 (0.02) & {\bf 0.71 (0.01)} \\
1,000 & Non-Linear & 25\% & 2.0 &      0.59 (0.02)  &      0.59 (0.02)  & 0.53 (0.02) & 0.51 (0.02) & 0.46 (0.03) & {\bf 0.69 (0.01)} \\
5,000 & Non-Linear & 25\% & 2.0 &      0.63 (0.01)  &      0.60 (0.01)  & 0.53 (0.01) & 0.51 (0.01) & 0.46 (0.02) & {\bf 0.68 (0.01)} \\
\midrule
1,000 & Linear     & 50\% & 0.5 &      0.64 (0.01)  &      0.64 (0.01)  & 0.56 (0.03) & 0.50 (0.02) & 0.38 (0.02) & {\bf 0.65 (0.01)} \\
5,000 & Linear     & 50\% & 0.5 & {\bf 0.67 (0.01)} &      0.65 (0.02)  & 0.56 (0.02) & 0.64 (0.01) & 0.39 (0.01) &      0.65 (0.01)  \\
1,000 & Linear     & 50\% & 2.0 &      0.62 (0.02)  &      0.62 (0.02)  & 0.53 (0.02) & 0.54 (0.02) & 0.44 (0.02) & {\bf 0.63 (0.01)} \\
5,000 & Linear     & 50\% & 2.0 &      0.59 (0.01)  & {\bf 0.63 (0.01)} & 0.54 (0.02) & 0.55 (0.01) & 0.44 (0.01) & {\bf 0.63 (0.02)} \\
1,000 & Non-Linear & 50\% & 0.5 &      0.58 (0.02)  &      0.62 (0.02)  & 0.60 (0.02) & 0.53 (0.02) & 0.45 (0.02) & {\bf 0.69 (0.01)} \\
5,000 & Non-Linear & 50\% & 0.5 &      0.59 (0.01)  &      0.61 (0.01)  & 0.60 (0.02) & 0.54 (0.01) & 0.44 (0.01) & {\bf 0.70 (0.01)} \\
1,000 & Non-Linear & 50\% & 2.0 &      0.62 (0.02)  &      0.60 (0.02)  & 0.62 (0.02) & 0.52 (0.02) & 0.46 (0.02) & {\bf 0.67 (0.01)} \\
5,000 & Non-Linear & 50\% & 2.0 &      0.61 (0.01)  &      0.61 (0.02)  & 0.62 (0.01) & 0.52 (0.01) & 0.45 (0.01) & {\bf 0.67 (0.01)} \\
\bottomrule
\end{tabular}
}
\end{table}

\subsection{Supplemental Simulation Studies}

\subsubsection{Bivariate Brier Score}

To examine the performance of the proposed bivariate Brier score, we generated 1,000 independent datasets of size $n = 1,000$ based on the illness-death model. Across all simulated datasets and simulation settings, we assumed Weibull baseline hazards with a shape parameter of 1.5 and a scale parameter of 0.2, and a population frailty variance of $\theta = 0.5$. We considered four simulation settings, varying whether the semi-competing outcomes depended on a uniform random covariate, and varying the administrative censoring rate at 0\% and 50\%. We calculated the integrated bivariate Brier score for 1-year survival over a grid of 100 evenly spaced time points, and compared the results from the model fit to a calculation which utilized the true model parameters. This comparison to the `truth' gives the degree of irreducible error in the bivariate Brier score for each setting. Table \ref{tab:sensitivity_bbs} shows that the results from the model fit were on par with those calculated using the true model parameters, giving an approximate lower bound for the bivariate Brier score.

\begin{table}[!ht]
\label{tab:bbs_sim}
\caption{Mean (SD) integrated bivariate Brier score (iBBS) under various data generation settings, averaged over 1,000 generated datasets for each setting to assess the degree of irreducible error in the iBBS.}
\label{tab:sensitivity_bbs}
\vspace{2ex}
\begin{tabular}{cccrr}
    \toprule
    Simulation Setting & Covariates Generated & Censoring Generated & True iBBS & Calculated iBBS \\
    \midrule
    \rowcolor{gray!10}
    1 & No  & No  & 0.0187 (0.0068) & 0.0199 (0.0073) \\
    2 & Yes & No  & 0.0181 (0.0067) & 0.0205 (0.0077) \\
    \rowcolor{gray!10}
    3 & No  & Yes & 0.0206 (0.0067) & 0.0219 (0.0072) \\
    4 & Yes & Yes & 0.0195 (0.0066) & 0.0221 (0.0075) \\
    \bottomrule
\end{tabular}
\end{table}

\subsubsection{Sensitivity Analysis - Gamma Frailty}
\label{sec:gamma_sensitivity}

\textcolor{black}{In a sensitivity analysis, we study the robustness of our method to the assumed gamma distribution of the latent frailties, $\gamma_i$, as this is the only parametric assumption we make in the proposed method. Specifically, we reproduce the results of our main analysis, where we generate $\gamma_i$ from a gamma distribution with mean 1 and variance $\theta$ and compare these results with settings where we instead generate $\gamma_i$ from a log-normal distribution with mean 1 and variance $\theta$, keeping all other aspects of the data generating mechanism fixed. For simplicity, we focus on one setting for the sample size $(n = 1,000)$ and one setting for the censoring rate (50\%), as these are the most realistic, and vary the true risk functions (linear versus nonlinear), the frailty variance ($\theta = 0.5$ versus 2.0) and the frailty distribution (gamma versus log-normal). The results of this analysis are given in Table \ref{tab:sensitivity_gamma}. Overall, these results show that our proposed approach performs well when the latent frailties truly follow a gamma distribution. When the latent frailties instead follow a log-normal distribution, we see that our approach is robust to this misspecification in terms of its predictive performance (measured by iBBS and C‐index), and its bias in estimating the baseline hazard functions for each state transition. However, we do note that the bias in the log-risk functions, $h_g$, grows noticeably with this misspecification, particularly when the frailty variance, $\theta$, is large or the true risk function is highly nonlinear.}

\begin{table}[!ht]
\tiny
\caption{\textcolor{black}{Sensitivity analysis comparing an assumed gamma versus log‐normal frailty: average (SD) mean integrated squared errors for each log‐risk function ($h_g(\boldsymbol{X}_i); g = 1,2,3$), integrated bivariate Brier score (iBBS), bivariate concordance (C) index, and average bias in estimating the baseline hazards under four simulation settings. All results are based on $n = 1,000$ observations and $50\%$ censoring.}}
\label{tab:sensitivity_gamma}
\centering
\vspace{2ex}
{\color{black}
\begin{tabular}{cccccccccc}
\toprule
\multicolumn{10}{c}{\bf Log-Risk Functions} \\
\midrule
\multicolumn{4}{c}{\bf Simulation Settings} & \multicolumn{2}{c}{\bf First Transition: $h_1(\boldsymbol{X}_i)$} & \multicolumn{2}{c}{\bf Second Transition: $h_2(\boldsymbol{X}_i)$} & \multicolumn{2}{c}{\bf Third Transition: $h_3(\boldsymbol{X}_i)$} \\
\cmidrule(lr){1-4} \cmidrule(lr){5-6} \cmidrule(lr){7-8} \cmidrule(lr){9-10}
Sample Size & Risk Function & Censoring Rate & $\theta$ & Gamma & Log-Normal & Gamma & Log-Normal & Gamma & Log-Normal \\
\midrule
1,000       & Linear        & 50\%	         & 0.5      & 0.73 (0.17) &  2.14 (0.25) & 0.88 (0.16) &  2.91 (0.48) & 0.43 (0.18) & 2.16 (0.25) \\
1,000       & Linear        & 50\%	         & 2.0      & 0.51 (0.17) &  4.80 (0.45) & 0.35 (0.11) &  6.14 (0.62) & 0.61 (0.18) & 6.41 (1.04) \\
1,000       & Non-Linear    & 50\%	         & 0.5      & 4.07 (0.52) &  5.46 (0.51) & 4.28 (0.55) &  5.76 (0.55) & 4.39 (0.58) & 5.27 (0.47) \\
1,000       & Non-Linear    & 50\%	         & 2.0      & 4.15 (0.48) & 11.14 (1.07) & 4.03 (0.58) & 13.52 (1.37) & 4.47 (0.54) & 10.89 (1.33) \\
\midrule
\multicolumn{10}{c}{\bf Performance Metrics and Baseline Hazards} \\
\midrule
\multicolumn{4}{c}{\bf Simulation Settings} & \multicolumn{2}{c}{\bf iBBS} & \multicolumn{2}{c}{\bf C-Index} & \multicolumn{2}{c}{$\boldsymbol{\lambda_{0g}(t)}$} \\
\cmidrule(lr){1-4} \cmidrule(lr){5-6} \cmidrule(lr){7-8} \cmidrule(lr){9-10}
Sample Size & Risk Function & Censoring Rate & $\theta$ & Gamma & Log-Normal & Gamma & Log-Normal & Gamma & Log-Normal \\
\midrule
1,000       & Linear     & 50\%	& 0.5 & 0.03 (0.002) & 0.03 (0.006) & 0.65 (0.01) & 0.64 (0.01) & 0.95 (1.49) & 0.95 (0.88) \\
1,000       & Linear     & 50\%	& 2.0 & 0.04 (0.006) & 0.07 (0.006) & 0.63 (0.01) & 0.65 (0.01) & 1.12 (2.48) & 0.47 (1.08) \\
1,000       & Non-Linear & 50\%	& 0.5 & 0.14 (0.004) & 0.05 (0.006) & 0.69 (0.01) & 0.60 (0.02) & 0.31 (0.27) & 0.77 (0.53) \\
1,000       & Non-Linear & 50\%	& 2.0 & 0.13 (0.010) & 0.09 (0.005) & 0.67 (0.01) & 0.69 (0.01) & 0.27 (0.45) & 0.53 (0.55) \\
\bottomrule
\end{tabular}
}
\end{table}

%--- SECTION 6 ------------------------------------------------------------------

\section{Boston Lung Cancer Study}
\label{sec:6}

\textcolor{black}{The motivation for this work comes from the Boston Lung Cancer Study (BLCS), a longitudinal epidemiological cancer cohort study. Patients are recruited on a rolling basis upon initial lung cancer diagnosis and followed until death. During the course of follow-up, disease recurrence and progression are recorded, which signify major non-terminal events in a patient's disease trajectory that modify their risk of mortality. For early-stage patients (Stages 1-3a), we define the non-terminal event time as time to first tumor recurrence, and for patients with late stage tumors, we define the non-terminal event time as the time to first progression. Disease recurrence and progression were determined by radiomic assessment based on the tumor growth, using combination of physical exam, imaging, and pathology based on the Response Evaluation Criteria in Solid Tumours (RECIST) guidelines \citep{eisenhauer2009new}.} All-cause mortality was reported to the BLCS and supplemented by the National Death Index and other sources. Our semi-competing events are cancer progression or death, which could be censored by the end of follow-up.

\subsection{Study Sample}

\noindent \textcolor{black}{Among the 19,497 participants in the BLCS, 7,755 were initially eligible for inclusion in this study. Initial eligibility was defined as having a positive lung cancer diagnosis. Participants were initially ineligible if they were enrolled with esophageal cancer or other primary cancer, no cancer upon further study, or as a negative control in the case of spouses, friends, or other participants. Among those 7,755 eligible patients, we identified 7,697 (99.3\%) with the temporal information necessary to define their semi-competing outcomes, namely (1) date of primary diagnosis, (2) recurrence, progression, and/or death date where applicable, and (3) last follow-up date or non-recurrence/progression date. Specifically, we removed 58 (0.7\%) patients without a diagnosis date. We further removed 56 patients (0.7\%) without adequate follow-up time, defined as a diagnosis date within six months of the study end date. Additionally, as small-cell lung cancer (SCLC) exhibits a distinct neuroendocrine biology, rapid doubling time, and unique chemotherapy-radiotherapy regimens compared to non-small-cell lung cancer (NSCLC), we removed 207 patients (2.7\%) with small-cell lung cancer, as well as six patients (0.08\%) with carcinoma \textit{in situ}, i.e., stage 0. Lastly, we removed 25 (0.3\%) patients whose death and/or progression dates coincided with their diagnosis dates. Our final analytic cohort consisted of $n$ = 7,403 (95.5\%) patients diagnosed with NSCLC between June 1983 and February 2023. Disease progression was reported in 2,443 (33.0\%) patients, with 1,570 (21.2\%) patients experiencing progression followed by death and 3,636 (49.1\%) patients who died prior to progression (see Table \ref{tab:events}).}

\begin{table} 
\centering
\caption{Semi-competing event rates among $n = 7,403$ patients in our analytic sample.}
\label{tab:events}
\vspace{2ex}
{\color{black}
\begin{tabular}{ccc}
    \toprule
    Progression Observed / Death Observed  &          Yes &           No \\
    \midrule
    Yes                  &    1,570 (21.21\%) &   873  (11.79\%) \\
    \rowcolor{gray!10}
    No                   & 3,636 (49.12\%) & 1,324 (17.88\%) \\ 
    \bottomrule
\end{tabular}
}
\end{table}

\textcolor{black}{Detailed information on patient demographics, smoking history, and physiologic measurements were  collected through questionnaire when the patient was recruited to the Boston Lung Cancer Study, at their time of diagnosis. Genetic mutations were also collected.} Potential demographic predictors included patient age at diagnosis (years), sex assigned at birth, self-identified race, and ethnicity. Smoking status and pack-years of smoking were also included. Relevant clinical predictors included cancer stage at diagnosis, initial treatment, indications of chronic obstructive pulmonary disease (COPD) or asthma, and oncogenic (somatic driver) mutation status (EGFR or KRAS). Table \ref{tab:desc} reports summary statistics for these risk factors in our study sample. \textcolor{black}{As shown, median (interquartile range; IQR) age at diagnosis was 67 (59, 74) years, with 3,966 (54\%) patients being female, 6,834 (92\%) being White, and 6,410 (87\%) being non-Hispanic. Clinically relevant features are as follows. The majority of patients had a history of smoking (6,259; 85\%), with a median (IQR) of 36 (11, 57) pack-years of smoking. Further, 1,553 patients (21\%) were tested using the SNaPshot assay for the presence of genetic variants. The results of this testing revealed that 405 (5.5\%) patients were positive for at least one KRAS variant and 298 (4.0\%) patients were positive for at least one EGFR variant. COPD was present in 2,284 (55\%) patients and 410 (7.7\%) patients had asthma. Lastly, 4,444 (60\%) patients initially underwent surgery, while 1,851 (25\%) patients initially received chemotherapy, 366 (4.9\%) received radiation, and 742 (10\%) received another form of treatment (Table \ref{tab:desc}). The distributions of these characteristics are similar to a recent study utilizing patient data from Massachusetts General Hospital, which draws comparisons to the BLCS cohort \citep{yuan2021performance}.}

\begin{table} 
    \scriptsize
    \caption{\textcolor{black}{Characteristics of the $n = 7,403$ patients with non-small cell lung cancer diagnosed between June 1983 and February 2023 in our analytic sample derived from the Boston Lung Cancer Study cohort. Summary statistics are reported as $n (\%)$ for categorical predictors and median (interquartile range) for continuous covariates.}}
    \label{tab:desc} 
    \vspace{4ex}
    {\color{black}
    \begin{tabular}{lcccc}
    \toprule
    & \textbf{Total} & \textbf{Cancer Stage: Early} & \textbf{Cancer Stage: Late} & \textbf{Cancer Stage: Unknown}\\
    \textbf{Characteristic} & \textbf{N = 7,403}\textsuperscript{1} & \textbf{N = 4,700} & \textbf{N = 2,342}& \textbf{N = 361}\\ 
    \midrule
    \rowcolor{gray!10}
    Age at Diagnosis (yrs.) & 67 (59, 74) & 68 (61, 75) & 64 (56, 72) & 66 (58, 73)\\
    \quad Unknown      & 220 & 212	& 7	& 1        \\
    \rowcolor{gray!10}
    Sex & & & & \\ 
    \quad Female  & 3,966 (54\%)  & 2,603 (55\%) & 1,188 (51\%) & 175 (48\%)\\
    \quad Male    & 3,431 (46\%) & 2,093 (45\%) & 1,152 (49\%) & 186 (52\%)\\ 
    \quad Unknown &   6 & 4 & 2 & 0\\
    \rowcolor{gray!10}
    Race & & & &\\
    \quad White/Caucasian     & 6,834 (92\%) & 4,349 (93\%) & 2,149 (92\%) & 336 (93\%) \\
    \quad Black/African American & 126 (1.7\%) & 83 (1.7\%) & 40 (1.7\%)&  3 (0.8\%)\\
    \quad Asian     &   140 (1.9\%) & 69 (1.5\%) & 67 (2.9\%) & 4 (1.1\%)\\
    \quad Other     &   98 (1.3\%) & 60 (1.3\%) & 35 (1.5\%) & 3 (0.8\%)\\
    \quad Unknown   &   205 (2.7\%) & 139 (3.0\%) & 51 (2.2\%)& 15 (4.1\%)\\
    \rowcolor{gray!10}
    Ethnicity & & & & \\
    \quad Non-Hispanic & 6,410 (87\%) & 3,990 (85\%) & 2,112 (90\%) & 308 (85\%)\\
    \quad Hispanic     &  87 (1.2\%) & 57 (1.2\%) & 28 (1.2\%)	& 2 (0.6\%) \\
    \quad Unknown      & 906 (13\%)  & 653 (14\%) & 202 (8.6\%) & 51 (14\%)\\
    % Education &  \\
    % \quad Some Grade School                  &   447 (6.0\%) \\
    % \quad Some High School                   &   991 (13\%)  \\
    % \quad High School Graduate               & 1,475 (20\%)  \\
    % \quad Vocational/Technical School        &   283 (3.8\%) \\    
    % \quad Some College or Associate's Degree & 1,487 (20\%)  \\
    % \quad College Graduate                   &   977 (13\%)  \\
    % \quad Graduate or Professional School    &   817 (11\%)  \\
    % \quad Other/Unknown                      &   983 (13\%)  \\
    \rowcolor{gray!10}
    Smoking Status & & & & \\
    \quad Smoker     & 6,259 (85\%) & 4,007 (85\%) & 1,917 (82\%) & 335 (93\%)\\
    \quad Non-Smoker &   1,009 (14\%) & 592 (13\%) & 402 (17\%) & 15 (4.2\%)\\
    \quad Unknown    &   135 (1.8\%) & 101 (2.1\%) & 23 (1.0\%) & 11 (3.0\%)\\
    \rowcolor{gray!10}
    Pack-Years of Smoking & 36 (11, 57) & 37 (12, 58) & 32 (8, 52) & 49 (30, 76)\\
    \quad Unknown    & 958 & 818 & 99 & 41\\
    %\rowcolor{gray!10}
    %Histologic Type &  \\
    %\quad Adenocarcinoma & 3,958 (53\%) \\
    %\quad Squamous Cell Carcinoma & 1,175 (16\%) \\
    %\quad Non-Small Cell Lung Cancer, Unspecified & 870 (12\%) \\
    %\quad Small Cell Lung Cancer & 301 (4.0\%) \\
    %\quad Other/Unknown & 1,156 (15\%) \\
    %\rowcolor{gray!10}
    \rowcolor{gray!10}
    Initial Treatment &  & & & \\
    \quad Surgery & 4,444 (60\%) & 3,994 (85\%) & 378 (16\%) & 72 (20\%)\\
    \quad Chemotherapy & 1,851 (25\%) & 365 (7.8\%) & 1,473 (62\%) & 13 (3.6\%) \\
    \quad Radiation & 366 (4.9\%) & 194 (4.1\%) & 163 (6.9\%) & 9 (2.4\%)\\
    \quad Other/Unknown & 742 (10\%) & 147 (3.1\%) & 328 (14\%) & 267 (74\%)\\
    \rowcolor{gray!10}
    EGFR Status & & & &  \\
    \quad Variant Negative & 1,255 (17\%) & 737 (16\%) & 498 (21\%) & 20 (5.5\%)\\
    \quad Variant Positive & 298 (4.0\%) & 158 (3.4\%) & 140 (6.0\%) & 0 (0\%)\\
    \quad Not Tested & 5,850 (79\%) & 3,805 (81\%) & 1,704 (73\%) & 341 (94\%)\\
    \rowcolor{gray!10}
    KRAS Status & & & & \\
    \quad Variant Negative & 1,148 (16\%) & 630 (13\%) & 500 (21\%) & 18 (5.0\%)\\
    \quad Variant Positive & 405 (5.5\%) & 265 (5.6\%) & 138 (5.9\%) & 2 (0.6\%)\\
    \quad Not Tested & 5,850 (79\%) & 3,805 (81\%) & 1,704 (73\%) & 341 (94\%) \\
    \rowcolor{gray!10}
    COPD\textsuperscript{2} & 2,284 (55\%) & 1,662 (61\%) & 505 (40\%) & 117 (54\%)\\
    Asthma & 410 (7.7\%) & 254 (7.6\%) & 136 (7.9\%) & 20 (7.1\%) \\
    \bottomrule
    \end{tabular}
    \textsuperscript{1}Median (IQR); n (\%) \textsuperscript{2}COPD: Chronic Obstructive Pulmonary Disease;
}
\end{table}

\subsection{Predictive Modeling}

\noindent \textcolor{black}{We applied our proposed method alongside those of \citet{xu2010statistical}, \citet{lee2015bayesian}, \citet{lee2017accelerated}, \citet{gorfine2021marginalized}, and \citet{kats2022accelerated} to estimate the hazards of cancer progression, mortality, and mortality following progression with the predictors listed in Table~\ref{tab:desc}. However, the methods of \citet{gorfine2021marginalized} and \citet{kats2022accelerated} did not  converge on this dataset, and their results are therefore omitted.} We used five-fold cross-validation to assess the predictive performance. That is, we randomly partitioned the samples into five folds. Each time, we trained the model using four of them (80\% samples), and reserved the rest (20\% samples) for validation. \textcolor{black}{Hyperparameters, including the number of nodes per hidden layer, the learning rate, the dropout rate, and the regularization rate, were optimized over a grid of candidate values and chosen based on best predictive performance.
Since the three-layer subnetworks worked well in simulations, we used a neural network with three hidden layers in each subnetwork. We then tuned the number of nodes per layer from a range of 16 to 1024 nodes. For this data analysis, the optimal number of nodes was 1024, 64, and 32 in the first, second, and third hidden layers, respectively. We further selected an optimal dropout rate of 0.1 and tuned the learning rate from 0.0001 to 0.05, with 0.0001 selected for the final analysis.}  We then tested the model performance on the remaining fold (20\% of patients) \textcolor{black}{and calculated the bivariate Brier score at one hundred evenly spaced time points (from time zero to five years post-diagnosis) and the bivariate C-index.} We repeated the procedure five times until each of the five folds was used for validation, and computed the average bivariate Brier score and bivariate C-index and their standard deviations. \textcolor{black}{Table \ref{tab:overall_performance_compare} compared our approach to the competing methods in these two metrics, showing that the proposed method demonstrates better predictive performance, with a lower bivariate Brier score and higher C-index compared to those obtained by \citet{lee2015bayesian} and \citet{lee2017accelerated}. While the method of \citet{xu2010statistical} shows somewhat better performance in terms of the bivariate Brier score, and comparable performance in terms of bivariate C-index, the proposed approach offers improved flexibility for capturing nonlinear effects and maintains competitive accuracy. }

\begin{table}[!ht]
    \centering
    \caption{\textcolor{black}{Overall Brier score (BBS) and concordance index (C-index) for each method. Lower BBS and higher C-index indicate better predictive performance.}}
    \label{tab:overall_performance_compare}
    {\color{black}
    \begin{tabular}{lcc}
    \toprule
    \textbf{Method} & \textbf{Brier Score (BBS)} & \textbf{C-index} \\
    \midrule
    Xu (2010)        & 0.18 (0.17-0.18) & 0.64 (0.61-0.67)\\
    \rowcolor{gray!10}
    Lee (2015)       & 0.53 (0.52-0.53) &  0.58 (0.54-0.61)\\
    Lee (2017)       & 0.53 (0.52-0.54) &  0.57 (0.53-0.60)\\
    \rowcolor{gray!10}
    Proposed Method  & 0.30 (0.29-0.31) &  0.64 (0.60-0.68)\\
    \bottomrule
    \end{tabular}
    }
\end{table}

\begin{figure}
    \includegraphics[width = 5.5in]{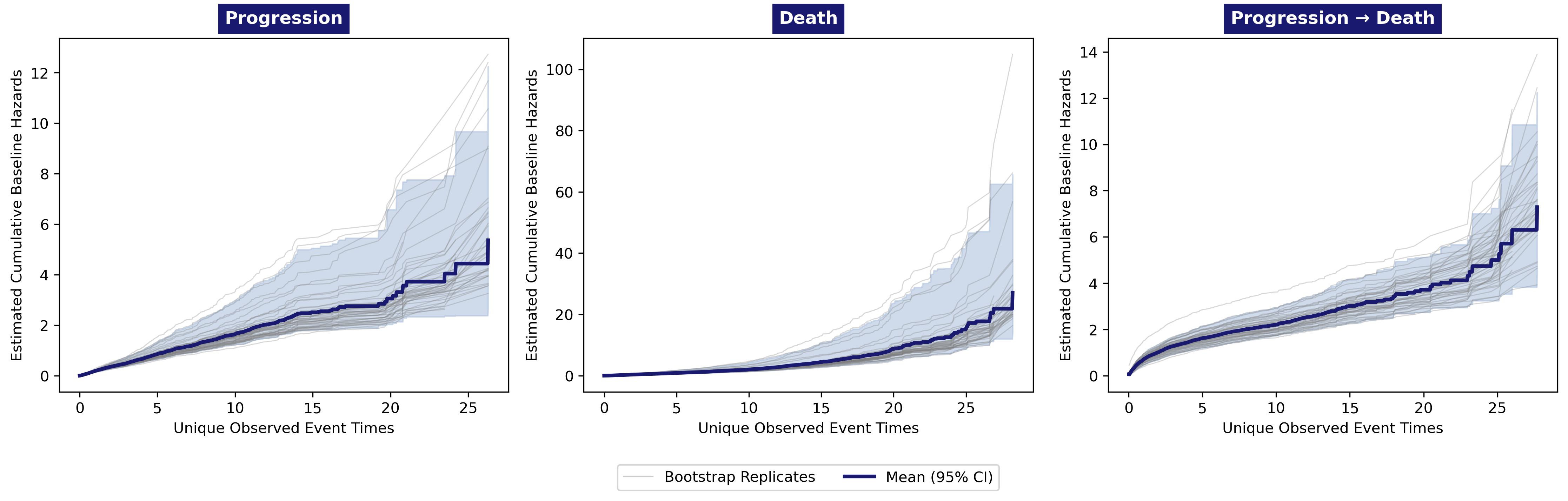}
    \caption{\textcolor{black}{Average estimated cumulative baseline hazard functions (solid blue lines) and 95\% bootstrap confidence intervals (blue bands) for each state transition based on 50 bootstrap samples of our data (solid gray lines).}}
    \label{fig:blcs_baseline}
\end{figure}

\textcolor{black}{We then applied the proposed neural EM algorithm to estimate the frailty variance, $\theta$, obtaining a value of 2.09. For context, this is approximately equal to a Kendall's $\tau$ value of 0.511 \citep{austin2017tutorial}. To quantify the associated uncertainty, we computed a bootstrap standard error of 0.04 based on 50 resamplings of the data with replacement. This nonzero estimate of frailty suggests the presence of moderate subject-level dependence across the three transitions. Figure~\ref{fig:blcs_baseline} displays the average estimated cumulative baseline hazard functions along with 95\% bootstrap confidence intervals constructed from these 50 replicates. As shown, the baseline hazards are highest between progression and death.}

\textcolor{black}{We then compare the results from the existing approaches in modeling the effect of each risk factor across the three state transitions in Table \ref{tab:blcs_results_compare}. Of note, the estimated log hazard ratios (log-HR) disagree in both sign and magnitude, suggesting conflicting results when specifying linear risk functions. Specifically, we focus on three particular risk factors to exemplify this: age at diagnosis, sex, and smoking status. While age at diagnosis is a significant risk factor for all three state transitions (progression, death, and death following progression) based on the methods of \citeauthor{xu2010statistical} and \citeauthor{lee2015bayesian}, we fail to detect a significant association based on the method of \citeauthor{lee2017accelerated}. Moreover, for progression, while the results based on \citeauthor{xu2010statistical}'s method show a small, positive association (log-HR: 0.01; SE: 0.002), indicating that older age is associated with a higher risk of progression, the results based on the method of \citeauthor{lee2015bayesian} show a small, negative association (log-HR: -0.04; SE: 0.002), suggesting that older age is protective. Similarly, females have significantly lower risks for all three state transitions (log-HR for progression: -0.19; SE: 0.05, log-HR for death: -0.39; SE: 0.05, and log-HR for death following progression: -0.32; SE: 0.07, respectively) based on the method of \citeauthor{xu2010statistical}, while we only detect a significant protective effect via \citet{lee2015bayesian} in the transition from progression to death (log-HR: -0.37; SE: 0.10), while the results using \citet{lee2017accelerated} show a significantly higher risk of death from diagnosis for females (log-HR: 0.79; SE: 0.11). Lastly, while we see a significantly higher risk of death following progression for current versus never smokers using all three methods (log-HR: 0.57; SE: 0.13, log-HR: 0.57; SE: 0.09, log-HR: 0.80; SE: 0.14, respectively), only \citeauthor{xu2010statistical}'s method finds significant risks for progression (log-HR: 0.40; SE: 0.10) and death following diagnosis (log-HR: 0.50; SE: 0.09) when comparing current versus never smokers. In contrast, the method of \cite{lee2015bayesian} finds a significant protective effect for the progression transition (log-HR: -0.23; SE: 0.08), while \citeauthor{lee2017accelerated}'s method finds a significant protective effect for the transition to death following diagnosis (log-HR: -0.23; SE: 0.09). Similar differences in the sign and magnitude of risk for the other predictors considered can be seen in Table \ref{tab:blcs_results_compare}.}

\begin{figure} 
 
    \includegraphics[width = 5.5in]{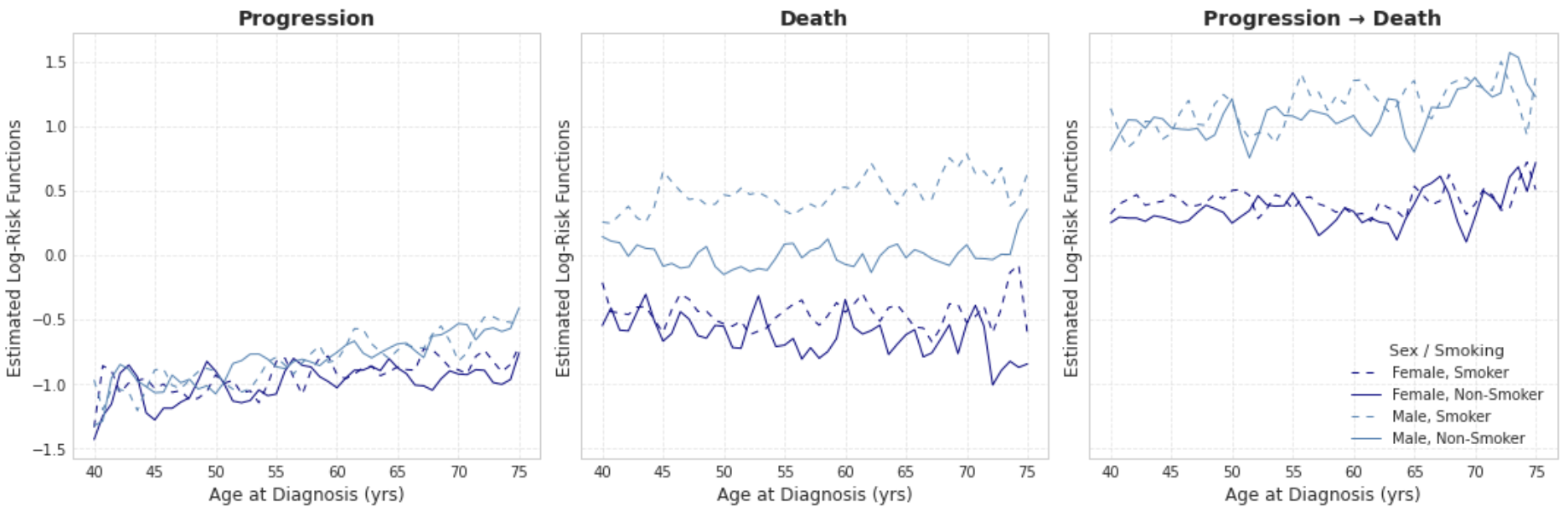}
    \caption{\textcolor{black}{Example nonparametric log-risk functions for age at diagnosis on each state transition, stratified by sex (line color) and smoking status (solid versus dashed lines). The predicted log risk functions for the proposed method were produced over a sequence of age (40 to 75 years), sex (male versus female), and smoking status (smoker versus non-smoker) values, holding all other predictors at their respective mean or modal values.}}
    \label{fig:blcs_log_risk}
\end{figure}

\textcolor{black}{To compare the results in Table \ref{tab:blcs_results_compare} to those from our method, Figure \ref{fig:blcs_log_risk} depicts the log-risk $(h)$ functions for the predicted effect of patient age at diagnosis on each state transition, take over a sequence of potential ages (40 to 75 years) and stratified by sex (male versus female) and smoking status (smoker versus non-smoker). All other covariates were fixed to be at their sample means or modes for illustration. As shown, there is a slight, increasing relationship between age and all three state transitions, but particularly in the transition from progression to death. Further, these relationships differ by sex, with males 65 and older having a greater increase in the risk of progression and death following progression. Smoking status appears to have a stronger effect on the risk of death from diagnosis, and to a lesser extent for the other state transitions, with male smokers having a higher risk of death than male non-smokers. Male patients are shown to have a higher risk of mortality than female patients, following both diagnosis and progression, and regardless of age and smoking status. However, the separation in risk between male smokers versus non-smokers and male versus female patients in the risk of death following diagnosis suggest potential interaction effects for this state transition.}

{\color{black}
\begin{longtable}{lrrr}
\caption{Log hazard ratio estimates and standard errors for three transition types.} \\
\label{tab:blcs_results_compare} \\
\toprule
{\bf Characteristic} & Xu (2010) & Lee (2015) & Lee (2017) \\
\midrule
\multicolumn{4}{l}{\bf Progression} \\
\midrule 
Age at Diagnosis (yrs.)                         &  0.01 (0.002)  & -0.04 (0.002) &  0.02 (0.02) \\
\rowcolor{gray!10}
Female (vs.~Male)                               & -0.19 (0.05)   & -0.03 (0.03)  &  0.07 (0.08) \\
Unknown Sex (vs.~Male)                          &  0.18 (0.80)   &  0.90 (0.71)  & -0.26 (0.19) \\
\rowcolor{gray!10}
Other Race (vs.~White/Caucasian)                &  0.45 (0.20)   &  0.42 (0.14)  & -0.55 (0.13) \\
Asian (vs.~White/Caucasian)                     &  0.05 (0.16)   &  0.35 (0.28)  &  0.11 (0.62) \\
\rowcolor{gray!10}
Black/African American (vs.~White/Caucasian)    &  0.15 (0.18)   & -0.01 (0.22)  &  0.06 (0.13) \\
Unknown Race (vs.~White/Caucasian)              & -0.43 (0.18)   &  0.09 (0.16)  &  0.18 (0.15) \\
\rowcolor{gray!10}
Hispanic (vs.~Non-Hispanic)                     &  0.36 (0.21)   &  0.02 (0.42)  & -0.15 (0.17) \\
Unknown Ethnicity (vs.~Non-Hispanic)            &  0.21 (0.07)   & -0.03 (0.05)  & -0.51 (0.13) \\
\rowcolor{gray!10}
COPD$^1$                                        &  0.33 (0.07)   &  0.22 (0.14)  & -0.50 (0.24) \\
Unknown COPD Status$^1$                         & -0.85 (0.10)   & -0.22 (0.12)  &  0.17 (0.14) \\
\rowcolor{gray!10}
Asthma                                          &  0.10 (0.11)   & -0.29 (0.19)  &  0.21 (0.27) \\
Unknown Asthma Status                           &  1.43 (0.10)   &  0.65 (0.29)  & -0.57 (0.13) \\
\rowcolor{gray!10}
Late Stage (3B-4, vs.~Early Stage 1-3A)         &  0.52 (0.07)   &  0.08 (0.14)  & -0.33 (0.12) \\
Radiation (vs.~Chemotherapy)                    & -0.17 (0.12)   &  0.32 (0.18)  & -0.40 (0.07) \\
\rowcolor{gray!10}
Surgery (vs.~Chemotherapy)                      & -1.08 (0.08)   & -0.14 (0.04)  &  0.35 (0.16) \\
Other First-Line Treatment (vs.~Chemotherapy)   &  0.36 (0.56)   & -0.004 (0.00) & -0.16 (0.27) \\
\rowcolor{gray!10}
Unknown First-Line Treatment (vs.~Chemotherapy) & -1.70 (0.16)   & -0.98 (0.60)  &  0.58 (0.14) \\
Former Smoker (vs.~Never Smoker)                &  0.22 (0.08)   &  0.06 (0.02)  & -0.23 (0.19) \\
\rowcolor{gray!10}
Current Smoker (vs.~Never Smoker)               &  0.40 (0.10)   & -0.23 (0.08)  & -0.30 (0.34) \\
Smoker, Status Unknown (vs.~Never Smoker)       &  0.14 (0.22)   &  0.01 (0.07)  & -0.13 (0.20) \\
\rowcolor{gray!10}
Pack-Years of Smoking                           & -0.003 (0.001) &  0.02 (0.004) &  0.03 (0.03) \\
EGFR Mutation                                   & -0.19 (0.11)   &  0.25 (0.13)  &  0.06 (0.67) \\
\rowcolor{gray!10}
KRAS Mutation                                   &  0.30 (0.10)   &  0.90 (0.11)  & -0.43 (0.24) \\
No Genetic Testing                              & -0.49 (0.07)   & -0.18 (0.15)  &  0.47 (0.28) \\
\midrule
\multicolumn{4}{l}{\bf Death} \\
\midrule
Age at Diagnosis (yrs.)                         &  0.03 (0.002)  &  0.03 (0.00)  & -0.01 (0.01)  \\
\rowcolor{gray!10}
Female (vs.~Male)                               & -0.39 (0.05)   &  0.27 (0.18)  &  0.79 (0.11)  \\
Unknown Sex (vs.~Male)                          & -0.63 (1.48)   & -0.02 (1.10)  & -0.06 (0.38)  \\
\rowcolor{gray!10}
Other Race (vs.~White/Caucasian)                &  0.32 (0.22)   &  0.16 (0.20)  &  0.31 (0.17)  \\
Asian (vs.~White/Caucasian)                     & -0.18 (0.19)   &  0.07 (0.27)  &  0.69 (0.17)  \\
\rowcolor{gray!10}
Black/African American (vs.~White/Caucasian)    &  0.14 (0.17)   &  0.08 (0.001) &  0.50 (0.28)  \\
Unknown Race (vs.~White/Caucasian)              & -0.60 (0.20)   &  0.02 (0.23)  &  0.31 (0.12)  \\
\rowcolor{gray!10}
Hispanic (vs.~Non-Hispanic)                     &  0.30 (0.25)   &  0.23 (0.30)  &  0.31 (0.33)  \\
Unknown Ethnicity (vs.~Non-Hispanic)            &  0.23 (0.09)   & -0.27 (0.07)  &  0.25 (0.21)  \\
\rowcolor{gray!10}
COPD$^1$                                        & -0.12 (0.06)   & -0.57 (0.37)  &  0.15 (0.32)  \\
Unknown COPD Status$^1$                         &  0.19 (0.06)   & -0.05 (0.03)  &  0.30 (0.33)  \\
\rowcolor{gray!10}
Asthma                                          & -0.16 (0.10)   &  0.29 (0.23)  &  0.48 (0.19)  \\
Unknown Asthma Status                           & -0.41 (0.07)   & -0.19 (0.06)  &  0.67 (0.18)  \\
\rowcolor{gray!10}
Late Stage (3B-4, vs.~Early Stage 1-3A)         &  1.56 (0.07)   &  0.97 (0.31)  & -0.13 (0.20)  \\
Radiation (vs.~Chemotherapy)                    & -0.01 (0.11)   &  0.04 (0.15)  & -0.17 (0.14)  \\
\rowcolor{gray!10}
Surgery (vs.~Chemotherapy)                      & -1.11 (0.07)   & -0.16 (0.05)  &  0.66 (0.23)  \\
Other First-Line Treatment (vs.~Chemotherapy)   & -0.61 (0.95)   & -1.19 (2.31)  &  0.31 (0.56)  \\
\rowcolor{gray!10}
Unknown First-Line Treatment (vs.~Chemotherapy) &  0.39 (0.07)   &  1.56 (0.15)  & -0.61 (0.58)  \\
Former Smoker (vs.~Never Smoker)                &  0.23 (0.08)   & -0.06 (0.03)  &  0.10 (0.12)  \\
\rowcolor{gray!10}
Current Smoker (vs.~Never Smoker)               &  0.50 (0.09)   & -0.12 (0.13)  & -0.23 (0.09)  \\
Smoker, Status Unknown (vs.~Never Smoker)       &  0.23 (0.23)   & -0.09 (0.11)  &  0.12 (0.19)  \\
\rowcolor{gray!10}
Pack-Years of Smoking                           &  0.003 (0.001) &  0.07 (0.01)  &  0.02 (0.002) \\
EGFR Mutation                                   & -0.42 (0.16)   &  0.04 (0.27)  &  0.68 (0.21)  \\
\rowcolor{gray!10}
KRAS Mutation                                   &  0.10 (0.14)   &  0.01 (0.10)  &  0.50 (0.16)  \\
No Genetic Testing                              &  0.28 (0.08)   &  0.12 (0.10)  &  0.14 (0.41)  \\
\midrule
\multicolumn{4}{l}{\bf Progression $\to$ Death} \\
\midrule
Age at Diagnosis (yrs.)                         &  0.04 (0.003) &  0.03 (0.001) &  0.02 (0.01)  \\
\rowcolor{gray!10}
Female (vs.~Male)                               & -0.32 (0.07)  & -0.37 (0.10)  & -0.13 (0.12)  \\
Unknown Sex (vs.~Male)                          & -0.58 (1.29)  &  0.68 (0.73)  &  0.33 (0.28)  \\
\rowcolor{gray!10}
Other Race (vs.~White/Caucasian)                & -0.16 (0.27)  & -0.24 (0.12)  & -0.004 (0.10) \\
Asian (vs.~White/Caucasian)                     & -0.23 (0.23)  & -0.27 (0.19)  & -0.02 (0.09)  \\
\rowcolor{gray!10}
Black/African American (vs.~White/Caucasian)    &  0.19 (0.23)  &  0.04 (0.08)  & -0.41 (0.10)  \\
Unknown Race (vs.~White/Caucasian)              & -0.27 (0.26)  &  0.06 (0.49)  &  0.21 (0.20)  \\
\rowcolor{gray!10}
Hispanic (vs.~Non-Hispanic)                     &  0.36 (0.21)  &  0.02 (0.42)  & -0.15 (0.17)  \\
Unknown Ethnicity (vs.~Non-Hispanic)            &  0.21 (0.07)  & -0.03 (0.05)  & -0.51 (0.13)  \\
\rowcolor{gray!10}
COPD$^1$                                        &  0.004 (0.10) &  0.18 (0.09)  &  0.49 (0.06)  \\
Unknown COPD Status$^1$                         &  0.33 (0.14)  & -0.02 (0.07)  &  0.22 (0.07)  \\
\rowcolor{gray!10}
Asthma                                          & -0.11 (0.16)  & -0.07 (0.08)  &  0.22 (0.11)  \\
Unknown Asthma Status                           & -0.58 (0.15)  & -0.34 (0.11)  &  0.14 (0.05)  \\
\rowcolor{gray!10}
Late Stage (3B-4, vs.~Early Stage 1-3A)         &  0.71 (0.09)  &  0.45 (0.33)  & -0.68 (0.13)  \\
Radiation (vs.~Chemotherapy)                    &  0.07 (0.14)  &  0.41 (0.45)  & -0.26 (0.12)  \\
\rowcolor{gray!10}
Surgery (vs.~Chemotherapy)                      & -1.16 (0.09)  & -0.44 (0.10)  &  0.69 (0.18)  \\
Other First-Line Treatment (vs.~Chemotherapy)   &  0.36 (0.56)  & -0.004 (0.00) & -0.16 (0.27)  \\
\rowcolor{gray!10}
Unknown First-Line Treatment (vs.~Chemotherapy) &  0.01 (0.21)  &  0.98 (0.39)  & -0.65 (0.24)  \\
Former Smoker (vs.~Never Smoker)                &  0.23 (0.11)  &  0.07 (0.11)  &  0.46 (0.19)  \\
\rowcolor{gray!10}
Current Smoker (vs.~Never Smoker)               &  0.57 (0.13)  &  0.57 (0.09)  &  0.80 (0.14)  \\
Smoker, Status Unknown (vs.~Never Smoker)       & -0.30 (0.30)  & -0.23 (0.38)  &  0.09 (0.28)  \\
\rowcolor{gray!10}
Pack-Years of Smoking                           &  0.01 (0.001) &  0.01 (0.003) & -0.06 (0.004) \\
EGFR Mutation                                   & -0.04 (0.15)  & -0.37 (0.02)  & -0.21 (0.12)  \\
\rowcolor{gray!10}
KRAS Mutation                                   & -0.02 (0.12)  & -0.01 (0.35)  &  0.14 (0.13)  \\
No Genetic Testing                              &  0.22 (0.09)  &  0.12 (0.03)  &  0.14 (0.12)  \\
\bottomrule
\end{longtable}
\vspace{-4ex}
\noindent $^1$COPD: Chronic Obstructive Pulmonary Disease
}

%--- SECTION 7 ------------------------------------------------------------------

\section{Discussion}
\label{sec:disc}

\noindent We propose a new neural expectation-maximization algorithm which, through a mixture of neural network architectures and trainable parameters, predicts time-to-event outcomes arising from a semi-competing risk framework (i.e., when a non-terminal event such as disease progression, modifies the risk of a patient's future survival). While previous work has developed machine learning approaches for multi-state or competing-risk settings \cite{tjandra2021hierarchical, lee2018deephit, lee2019dynamic, aastha2020deepcompete}, in which progression and death are considered independent of each other, we propose a new approach to further study the correlation between time to progression and death, and the modified hazards for mortality among those with progression and without progression. In simulation, our method could accurately estimate the relationship between covariates and the hazards of transitioning from disease onset to progression and death, particularly in situations where the risk relationship is complex.

\textcolor{black}{Our results detected several nonlinear effects and interactions between commonly-studied risk factors such as age, sex assigned at birth, and smoking status. As shown in the predicted risk functions for our data, we note a potential interaction between sex assigned at birth and smoking status in the hazards for mortality. Such findings have been corroborated in \cite{guo2009impact} and \cite{tseng2022association}. They found that patterns of smoking differed by other well-known risk factors such as patient age, sex, tumor stage, and histology, with smoking and tumor stage being predictive of patient mortality. Further, they found significant interactions between smoking, clinical stage, and age with respect to progression. Our study notes potential interactions between smoking status, age, and sex with respect to progression, as seen by the crossing of the covariate-specific risk functions. These crossings imply that the impact of smoking or sex is not uniform across age. For instance, being male or a smoker may confer higher risk primarily in older age, while differences are attenuated at younger ages. Such interactions are difficult to capture using traditional linear models but emerge naturally from our flexible, nonparametric approach. We also note that the difference in mortality hazards becomes more pronounced between males and females for each year higher age at diagnosis. The study by \cite{tseng2022association} examined the interacting effects between smoking status and other risk factors in patients with small-cell lung cancer, including age, sex, stage, and initial treatment. However, their study, which focused on small-cell lung cancer, found that non-smokers had higher hazards for mortality, while the opposite is true in our cohort of small-cell and non-small cell cases. As opposed to these works, which considered the end points of progression and death without their shared dependence, our analysis treats these outcomes as semi-competing. In applying our method to the prediction of semi-competing outcomes in the BLCS Cohort, we found that our neural EM approach had a greater predictive accuracy than some traditional semi-competing regression approaches. This is promising, as mortality is often assessed without the consideration of progression as a competing event, or to avoid technical difficulties such as dependent censoring, composite endpoints such as progression-free survival (which measure the time to the first of multiple possible events) are constructed. However, composite endpoints may mask the dependence of predictors on different endpoints, as the effects of certain clinical factors may differ across differing states in a patient's disease trajectory \citep{amir2012poor, chakravarty2008use}. Having a method which accounts for the dependence between multiple event types will help improve clinical decision-making.} 

It is also advantageous to use deep learning for risk prediction, as deep neural networks have the ability to accommodate potentially high-dimensional predictors. Recent works have shown that estimates based on multilayer feedforward neural networks are able to circumvent the {\it curse of dimensionality} in nonparametric settings \citep{bauer2019deep, poggio2017and}. Though fully understanding this phenomenon is a work in progress, several authors reason that neural networks project the data into a much lower relevant representational space through weighting \citep{abrol2021deep, goodfellow2016deep}. 

\textcolor{black}{There are several open directions of the future work. Firstly, we assume a parametric frailty distribution. In our sensitivity analysis, we see that our approach is robust to misspecification of the frailty distribution in terms of overall predictive accuracy and in estimating the baseline hazards, but sensitive in terms of estimating the log-risk functions. An alternative may be a fully nonparametric approach by specifying a frailty with a finite, but unknown, number of mixture components \citep{gasperoni2020non, chee2021semiparametric}. Second, while we focus on time-independent covariates, future work may consider time-varying predictors, which would increase the utility of the method. Third, while our use of the bootstrap provides a practical means to quantify uncertainty in our estimates, we acknowledge that it remains an ad hoc solution without formal guarantees in deep learning settings. Developing rigorous inferential theory for neural network-based estimators, particularly in semiparametric or nonparametric models, remains an open and promising area for future research.}  Lastly, while we focus on the joint distribution of the observed survival times for both event processes simultaneously, sometimes it can be of interest to study the marginal distribution of the non-terminal event (e.g., disease progression) while addressing the dependent censoring incurred by death, and one may directly deal with the marginal distribution of disease progression in the presence of mortality as a semi-competing event. We will address these problems elsewhere.

\newpage

%=== REFERENCES ===================================

\singlespacing

\bibliographystyle{plainnat}
\bibliography{ref}

\newpage

%=== APPENDICES ===================================

\begin{appendix}

\setcounter{figure}{0}
\setcounter{table}{0}
\setcounter{equation}{0}

\renewcommand\thefigure{\thesection.\arabic{figure}}
\renewcommand\thetable{\thesection.\arabic{table}} 
\renewcommand\theequation{\thesection.\arabic{equation}} 

%--- APPENDIX A -----------------------------------------------------------------

\section{Illness-Death Model}
\label{sec:a}

\subsection{Notation} 
\label{sec:a.1}

%Recall our notation in Section 2 of the main text. 
Let $T_{1}$ and $T_{2}$ denote the times to a non-terminal and terminal event, respectively. Let $\lambda_1(t_1)$ denote the hazard of the non-terminal event at time $t_1$, $\lambda_2(t_2)$ denote the hazard of the terminal event at $t_2$ without experiencing the non-terminal event, and $\lambda_3(t_2 \mid t_1)$ denote the hazard of the terminal event at $t_2$ given the non-terminal event happening at $t_1 \leq t_2$. These hazard rates, corresponding to the transitions between states are defined as

\begin{align}
    \lambda_1(t_1) &= \lim_{\Delta \rightarrow 0} \Pr[T_1 \in [t_1, t_1 + \Delta) \mid T_1 \geq t_1, T_2 \geq t_1] / \Delta; \label{eq:a:lam1} \\[2ex]
    \lambda_2(t_2) &= \lim_{\Delta \rightarrow 0} \Pr[T_2 \in [t_2, t_2 + \Delta) \mid T_1 \geq t_2, T_2 \geq t_2] / \Delta; \label{eq:a:lam2} \\[2ex]
    \lambda_3(t_2 \mid t_1) &= \begin{cases} \lim_{\Delta \rightarrow 0} \Pr[T_2 \in [t_2, t_2 + \Delta) \mid T_1 = t_1, T_2 \geq t_2] / \Delta, & t_2 > t_1 > 0; \\[1ex] 0, & {\rm otherwise}. \end{cases} \label{eq:a:lam3}
\end{align}

\noindent Note that the definitions of $\lambda_1(t_1)$ and $\lambda_2(t_2)$ mirror that of the cause-specific hazards under a competing risks framework, where they describe the hazards of first observing either the non-terminal or terminal event. Under semi-competing risks, observing the non-terminal event is subject to observing the terminal event, but not vice-versa. Hence, $\lambda_3(t_2\mid t_1)$ describes the hazards of observing the terminal event at $t_2$ after having observed the non-terminal event at $t_1$. As we cannot observe the non-terminal event after the terminal event has been observed, the space of $(T_1, T_2)$ is restricted to the so-called `upper wedge' of the first quadrant where $t_1 \leq t_2$, and the non-terminal event is said to be dependently censored by the terminal event. To incorporate this dependence, we parameterize (\ref{eq:a:lam1}) - (\ref{eq:a:lam3}) by extending the Cox proportional hazards model \citep{cox1972regression} to a shared gamma-frailty conditional Markov model, given by

\begin{align}
    \lambda_1(t_1 \mid \gamma) &= \gamma \lambda_{01}(t_1) \exp\{h_1(\boldsymbol{x})\}; \label{eq:a:condhaz1} \\[2ex]
    \lambda_2(t_2 \mid \gamma) &= \gamma \lambda_{02}(t_2) \exp\{h_2(\boldsymbol{x})\}; \label{eq:a:condhaz2} \\[2ex]
    \lambda_3(t_2 \mid t_1, \gamma) &= \begin{cases} \gamma \lambda_{03}(t_2) \exp\{h_3(\boldsymbol{x})\}, & t_2 > t_1 > 0; \\[1ex] 0, & {\rm otherwise}, \end{cases} \label{eq:a:condhaz3}
\end{align}

\noindent where $\gamma$ is a random effect, referred to as a subject's {\it frailty}, $\lambda_{0g}\left(t\right); g\in\{1,2,3\}$ are the baseline hazards for the three state transitions, $\boldsymbol{x}$ is a $p$-vector of covariates, and $h_g(\boldsymbol{x})$ are log-risk functions which relate the covariates to the hazard rates for each transition. Based on (\ref{eq:a:condhaz1}) - (\ref{eq:a:condhaz3}), survival functions of interest are

\begin{align}
    S(t_1, t_1 \mid \gamma) &= \Pr(T_1 > t_1, T_2 > t_1 \mid \gamma) = \exp\left\{-\gamma\left[\Lambda_{01}(t_1)\exp\left\{h_1(\boldsymbol{x})\right\} + \Lambda_{02}(t_1)\exp\left\{h_2(\boldsymbol{x})\right\}\right]\right\} \label{eq:a:st1t1}\\[2ex]
    S_{2 \mid 1}(t_2 \mid t_1, \gamma) &= \Pr(T_2 > t_2 \mid T_1 = t_1, \gamma) = \exp\left\{-\gamma[\Lambda_{03}(t_2) - \Lambda_{03}(t_1)]\exp\left\{h_3(\boldsymbol{x})\right\}\right\};\ t_2 > t_1 > 0, \label{eq:a:st2t1}
\end{align}

\noindent where $\Lambda_{0g}(t) = \int_0^t \lambda_{0g}(u)du$ are the cumulative baseline hazards for each transition. The joint survival function evaluated at $t_2 = t_1$ takes the form in (\ref{eq:a:st1t1}) due to the competing nature of the state transitions from no event to the first of either the non-terminal or the terminal event. The survival function of $t_2$ conditional on $t_1$ and the frailty in (\ref{eq:a:st2t1}) is defined in terms of the difference in time between events.

\subsection{Conditional Likelihood}
\label{sec:a.2}

\noindent As both events are subject to censoring, we do not fully observe $(T_1, T_2)$. With censoring time, $C$, we observe
\[
\mathcal{D} = \{(Y_{i1}, \delta_{i1}, Y_{i2}, \delta_{i2}, \boldsymbol{x}_i);\ i = 1, \ldots, n\},
\]

\noindent where $Y_{i2} = \min(T_{i2}, C_i)$, $\delta_{i2} = I(T_{i2} \leq C_i)$, $Y_{i1} = \min(T_{i1}, Y_{i2})$, $\delta_{i1} = I(T_{i1} \leq Y_{i2})$, $\boldsymbol{x}_i$ is a $p$-vector of covariates, $I(\cdot)$ is the indicator function, and $i$ indexes an individual in the study, $i = 1, \ldots, n$. There are four potential observation types for an individual during a finite period of follow-up, as given in Table \ref{tab:a1}.

\begin{table}[!ht]
    \centering
    \caption{{Potential event progressions and corresponding observed data}}
    \vspace{1ex}
    \label{tab:a1}
    \begin{tabular}{ccccccc}
        \toprule
        & \multicolumn{2}{c}{Observed Event Progression} & \multicolumn{4}{c}{Observed Data} \\
        \midrule
        Case & Non-Terminal & Terminal & $Y_{i1}$ & $\delta_{i1}$ & $Y_{i2}$ & $\delta_{i2}$ \\
        \midrule
        \rowcolor{gray!10}
        1 & \cmark & \cmark & $T_{i1}$ & 1 & $T_{i2}$ & 1 \\
        2 & \xmark & \cmark & $T_{i2}$ & 0 & $T_{i2}$ & 1 \\
        \rowcolor{gray!10}
        3 & \cmark & \xmark & $T_{i1}$ & 1 & $C_{i}$  & 0 \\
        4 & \xmark & \xmark & $C_{i}$  & 0 & $C_{i}$  & 0 \\
        \bottomrule
    \end{tabular}
\end{table}

\noindent To construct the likelihood conditional on the subject-specific frailties, we multiply the likelihood contributions under each case in Table \ref{tab:a1}, raised to the appropriate event indicators, and taken over the $n$ subjects. Define $\boldsymbol{\gamma} = (\gamma_1, \ldots, \gamma_n)$ to be the $n$-vector of latent frailties, and let $\boldsymbol{\psi} = \{\Lambda_{01}, \Lambda_{02}, \Lambda_{03}, \theta\}$ denote the collection of model parameters to be learned. The likelihood function is

\[
\begin{aligned}
    L&(\boldsymbol{\psi}; \mathcal{D}, \boldsymbol{\gamma}) = \prod_{i = 1}^{n} \left[\lambda_1(Y_{i1}) S(Y_{i1}, Y_{i1} \mid \gamma_i) \times \lambda_3(Y_{i2}) S_{2\mid 1}(Y_{i2} \mid Y_{i1}, \gamma_i)\right]^{\delta_{i1}\delta_{i2}} \times \left[\lambda_2(Y_{i2}) S(Y_{i1}, Y_{i1} \mid \gamma_i)\right]^{(1-\delta_{i1})\delta_{i2}} \\
    &\quad \quad \times \left[\lambda_1(Y_{i1}) S(Y_{i1}, Y_{i1} \mid \gamma_i) \times S_{2\mid 1}(Y_{i2}\mid Y_{i1}, \gamma_i)\right]^{\delta_{i1}(1-\delta_{i2})} \times \left[S(Y_{i1}, Y_{i1}\mid \gamma_i)\right]^{(1-\delta_{i1})(1-\delta_{i2})} \\[2ex]
    &= \prod_{i = 1}^{n} \left[\gamma_i \lambda_{01}(Y_{i1}) \exp\{h_1(\boldsymbol{x}_i)\} \exp\left\{-\gamma_i\left[\Lambda_{01}(Y_{i1})\exp\left\{h_1(\boldsymbol{x}_i)\right\} + \Lambda_{02}(Y_{i1})\exp\left\{h_2(\boldsymbol{x}_i)\right\}\right]\right\}\right. \\
    &\quad \quad \quad \quad \left.\times \gamma_i \lambda_{03}(Y_{i2}) \exp\{h_3(\boldsymbol{x}_i)\} \exp\left\{-\gamma_i[\Lambda_{03}(Y_{i2}) - \Lambda_{03}(Y_{i1})]\exp\left\{h_3(\boldsymbol{x}_i)\right\}\right\}\right]^{\delta_{i1}\delta_{i2}} \\
    &\quad \quad \times \left[\gamma_i \lambda_{02}(Y_{i2}) \exp\{h_2(\boldsymbol{x}_i)\} \exp\left\{-\gamma_i\left[\Lambda_{01}(Y_{i1})\exp\left\{h_1(\boldsymbol{x}_i)\right\} + \Lambda_{02}(Y_{i1})\exp\left\{h_2(\boldsymbol{x}_i)\right\}\right]\right\}\right]^{(1-\delta_{i1})\delta_{i2}} \\
    &\quad \quad \times \left[\gamma_i \lambda_{01}(Y_{i1}) \exp\{h_1(\boldsymbol{x}_i)\} \exp\left\{-\gamma_i\left[\Lambda_{01}(Y_{i1})\exp\left\{h_1(\boldsymbol{x}_i)\right\} + \Lambda_{02}(Y_{i1})\exp\left\{h_2(\boldsymbol{x}_i)\right\}\right]\right\}\right. \\
    &\quad \quad \quad \quad \left.\times \exp\left\{-\gamma_i[\Lambda_{03}(Y_{i2}) - \Lambda_{03}(Y_{i1})]\exp\left\{h_3(\boldsymbol{x}_i)\right\}\right\}\right]^{\delta_{i1}(1-\delta_{i2})} \\
    &\quad \quad \times \left[\exp\left\{-\gamma_i \left[\Lambda_{01}(Y_{i1}) \exp\{h_1(\boldsymbol{x}_i)\} + \Lambda_{02}(Y_{i1}) \exp\{h_2(\boldsymbol{x}_i)\}\right]\right\}\right]^{(1 - \delta_{i1})(1 - \delta_{i2})} \\[2ex]
    &= \prod_{i = 1}^{n} \left[\gamma_i^2 \lambda_{01}(Y_{i1}) \exp\{h_1(\boldsymbol{x}_i)\} \lambda_{03}(Y_{i2}) \exp\{h_3(\boldsymbol{x}_i)\} \times \exp\left\{-\gamma_i\left[\Lambda_{01}(Y_{i1})\exp\left\{h_1(\boldsymbol{x}_i)\right\}\right.\right.\right. \\
    &\quad \quad \quad \quad \left.\left.\left. +\ \Lambda_{02}(Y_{i1})\exp\left\{h_2(\boldsymbol{x}_i)\right\} + \left\{\Lambda_{03}(Y_{i2}) - \Lambda_{03}(Y_{i1})\right\}\exp\left\{h_3(\boldsymbol{x}_i)\right\}\right]\right\}\right]^{\delta_{i1}\delta_{i2}} \\
    &\quad \quad \times \left[\gamma_i \lambda_{02}(Y_{i2}) \exp\{h_2(\boldsymbol{x}_i)\} \times \exp\left\{-\gamma_i\left[\Lambda_{01}(Y_{i1})\exp\left\{h_1(\boldsymbol{x}_i)\right\} + \Lambda_{02}(Y_{i1})\exp\left\{h_2(\boldsymbol{x}_i)\right\}\right]\right\}\right]^{(1-\delta_{i1})\delta_{i2}} \\
    &\quad \quad \times \left[\gamma_i \lambda_{01}(Y_{i1}) \exp\{h_1(\boldsymbol{x}_i)\} \times \exp\left\{-\gamma_i\left[\Lambda_{01}(Y_{i1})\exp\left\{h_1(\boldsymbol{x}_i)\right\} + \Lambda_{02}(Y_{i1})\exp\left\{h_2(\boldsymbol{x}_i)\right\}\right.\right.\right. \\
    &\quad \quad \quad \quad \left.\left. +\ \left\{\Lambda_{03}(Y_{i2}) - \Lambda_{03}(Y_{i1})\right\}\exp\left\{h_3(\boldsymbol{x}_i)\right\}\right\}\right]^{\delta_{i1}(1-\delta_{i2})} \\
    &\quad \quad \times \left[\exp\left\{-\gamma_i \left[\Lambda_{01}(Y_{i1}) \exp\{h_1(\boldsymbol{x}_i)\} + \Lambda_{02}(Y_{i1}) \exp\{h_2(\boldsymbol{x}_i)\}\right]\right\}\right]^{(1 - \delta_{i1})(1 - \delta_{i2})}
\end{aligned}
\]

\[
\begin{aligned}
    &= \prod_{i = 1}^{n} \left[\gamma_i \lambda_{01}(Y_{i1}) \exp\{h_1(\boldsymbol{x}_i)\}\right]^{\delta_{i1}} \times \left[\gamma_i \lambda_{02}(Y_{i2}) \exp\{h_2(\boldsymbol{x}_i)\}\right]^{(1-\delta_{i1})\delta_{i2}} \times \left[\gamma_i \lambda_{03}(Y_{i2}) \exp\{h_3(\boldsymbol{x}_i)\}\right]^{\delta_{i1}\delta_{i2}} \\
    &\quad \quad \times \exp\left\{-\gamma_i \left[\Lambda_{01}(Y_{i1}) \exp\{h_1(\boldsymbol{x}_i)\} + \Lambda_{02}(Y_{i1}) \exp\{h_2(\boldsymbol{x}_i)\}\right]\right\}^{(1-\delta_{i1})} \\
    &\quad \quad \times \exp\left\{-\gamma_i \left[\Lambda_{01}(Y_{i1}) \exp\{h_1(\boldsymbol{x}_i)\} + \Lambda_{02}(Y_{i1}) \exp\{h_2(\boldsymbol{x}_i)\} + \left\{\Lambda_{03}(Y_{i2}) - \Lambda_{03}(Y_{i1})\right\} \exp\{h_3(\boldsymbol{x}_i)\}\right]\right\}^{\delta_{i1}}.
\end{aligned}
\]

\vspace{2ex}

\noindent Consolidating the remaining terms yields the final expression for the likelihood function, given by

\begin{align}
\begin{split}
\label{eq:a:condlik}
    L&(\boldsymbol{\psi}; \mathcal{D}, \boldsymbol{\gamma}) = \prod_{i = 1}^{n} \gamma_i^{\delta_{i 1}+\delta_{i 2}} \times \left[\lambda_{01}\left(Y_{i 1}\right) \exp\{h_1(\boldsymbol{x}_i)\}\right]^{\delta_{i 1}} \\
    &\times \left[\lambda_{02}\left(Y_{i 2}\right)\exp\{h_2(\boldsymbol{x}_i)\}\right]^{\left(1-\delta_{i 1}\right) \delta_{i 2}} \times \left[\lambda_{03}\left(Y_{i 2} \right)\exp\{h_3(\boldsymbol{x}_i)\}\right]^{\delta_{i 1} \delta_{i 2}} \\[2ex]
    & \times \exp\left\{-\gamma_i\left[\Lambda_{01}\left(Y_{i 1}\right)\exp\{h_1(\boldsymbol{x}_i)\} + \Lambda_{02}\left(Y_{i 1}\right)\exp\{h_2(\boldsymbol{x}_i)\} \right.\right. \\[2ex]
    &\quad \quad \left.\left. +\ \delta_{i 1}[\Lambda_{03}(Y_{i 2}) - \Lambda_{03}(Y_{i 1})] \exp\{h_3(\boldsymbol{x}_i)\}\right]\right\}.
\end{split}
\end{align}

\noindent As no maximizer exists for this function over the space of absolutely continuous cumulative baseline hazards, we constrain the parameter space of cumulative baseline hazards, $\Lambda_{01}$, $\Lambda_{02}$, and $\Lambda_{03}$, to consist of piecewise constant CADLAG (right continuous with left-hand limits) functions, with jumps occurring at observed event times. The maximizers over this discrete space are termed non-parametric maximum likelihood estimates (NPMLEs) of $\Lambda_{01}$, $\Lambda_{02}$, and $\Lambda_{03}$. In this framework, we modify the likelihood function (\ref{eq:a:condlik}) by replacing $\lambda_{0g}(t)$ with $\Delta\Lambda_{0g}(t)=\Lambda_{0g}(t)-\Lambda_{0g}(t-)$, representing the jump size at $t$. With this substitution, and multiplication by the density function for our frailty term, we obtain

\begin{align}
\begin{split}
\label{eq:a:condlik2}
    \tilde{L} &\left(\boldsymbol{\psi}; \mathcal{D}, \boldsymbol{\gamma}\right) = \prod_{i=1}^{n} \frac{\theta^{-\frac{1}{\theta}}}{\Gamma\left(\frac{1}{\theta}\right)} \times \gamma_{i}^{\frac{1}{\theta}-1} \times e^{-\frac{\gamma_{i}}{\theta}}\times \gamma_{i}^{\delta_{i 1}+\delta_{i 2}} \times \left[\Delta\Lambda_{01}\left(Y_{i1}\right) e^{h_1(\boldsymbol{x}_i)}\right]^{\delta_{i 1}} \\[1ex]
    & \times \left[\Delta\Lambda_{02}\left(Y_{i2}\right)e^{h_2(\boldsymbol{x}_i)}\right]^{\left(1-\delta_{i 1}\right) \delta_{i 2}} \times \left[\Delta\Lambda_{03}\left(Y_{i2} \right)e^{h_3(\boldsymbol{x}_i)}\right]^{\delta_{i 1} \delta_{i 2}}\\[1ex]
    &\times \exp\left\{-\gamma_{i}\left[\Lambda_{01}\left(Y_{i1}\right)e^{h_1(\boldsymbol{x}_i)} + \Lambda_{02}\left(Y_{i1}\right)e^{h_2(\boldsymbol{x}_i)}\right.\right. \\[1ex]
    &\quad \left.\left.+\ \delta_{i 1}[\Lambda_{03}(Y_{i2}) - \Lambda_{03}(Y_{i1})] e^{h_3(\boldsymbol{x}_i)}\right]\right\},
\end{split}
\end{align}

\vspace{1ex}

\noindent the `complete' likelihood (\ref{eq:a:condlik2}) defined on a step-wise constant space for  $\Lambda_{0g}, g=1,2,3,$. This  will serve as the basis of our proposed neural expectation-maximization (EM) algorithm, detailed below.

%--- APPENDIX B -----------------------------------------------------------------

\newpage

\setcounter{figure}{0}
\setcounter{table}{0}
\setcounter{equation}{0}

\section{Neural Expectation-Maximization Algorithm}
\label{sec:b}

In the following, we provide additional detail on the neural expectation-maximization (EM) algorithm outlined in Section 3 in the main text. Viewing the subject-specific frailties as missing data, the algorithm iterates between three steps, namely the expectation (E) step, the maximization (M) step, and the neural (N) step. In the E-step, we compute the conditional expectation of the log-likelihood (\ref{eq:a:condlik2}) given the observed data, $\mathcal{D}$, and the current estimates of $\boldsymbol{\psi}$, denoted by $\boldsymbol{\psi}_c$, wherein the conditional expectation is with respect to the distribution of $\gamma_i \mid \mathcal{D}, \boldsymbol{\psi}_c$. In the M-step, we calculate the NPMLEs of $\Lambda_{0g}$ while assuming the known values of $h_g$. Subsequently, we substitute these estimates into the conditional expectation of (\ref{eq:a:condlik2}). This process yields the conditional expectation of the log `profile' likelihood, serving as the objective function for utilizing deep neural networks to derive estimates for the log risk functions and frailty variance.

\subsection{Conditional Frailty Distributions} 
\label{sec:b.1}

We detail the derivations of the conditional densities of $\boldsymbol{\gamma}$ and $\log(\boldsymbol{\gamma})$ given the data, two quantities needed in the proposed Neural EM algorithm. Denote by {$\tilde{L}\left(\boldsymbol{\psi}; \mathcal{D}, \boldsymbol{\gamma}\right)$}, which was derived in Section \ref{sec:a.2}, the likelihood of the `complete' data, and by $f(\gamma_i)$ the marginal density of $\gamma_i$. We assume that, independent of $\boldsymbol{x}_i$, each $\gamma_i$ independently follows a Gamma distribution with a density function

\begin{equation}
\label{eq:gammadens}
    f(\gamma_i) = \frac{\theta^{-\frac{1}{\theta}}}{ \Gamma\left(\frac{1}{\theta}\right)}\gamma_i^{\frac{1}{\theta}-1}\exp\left\{-\frac{\gamma_i}{\theta}\right\}   
\end{equation}

\noindent so that $\mathbb{E}[\gamma_i] = 1$ and $\text{Var}(\gamma_i) = \theta$, and the marginal density of $\boldsymbol{\gamma}$ is the product over the $n$ independent $\gamma_i$ densities. Thus, for a fixed value of $\theta$, the `posterior' distribution of $\boldsymbol{\gamma}$ is

\[
\begin{aligned}
    f&(\boldsymbol{\gamma} \mid \mathcal{D}, \boldsymbol{\psi}) \propto f(\boldsymbol{\gamma})\times \tilde{L}\left(\boldsymbol{\psi}; \mathcal{D}, \boldsymbol{\gamma}\right) = \prod_{i = 1}^{n} \frac{\theta^{-\frac{1}{\theta}}}{\Gamma\left(\frac{1}{\theta}\right)} \gamma_i^{\frac{1}{\theta}-1} \exp\left\{-\frac{\gamma_i}{\theta}\right\} \times \gamma_i^{\delta_{i 1}+\delta_{i 2}} \\
    & \times \left[\Delta\Lambda_{01}\left(Y_{i 1}\right) \exp\{h_1(\boldsymbol{x}_i)\}\right]^{\delta_{i 1}} \times \left[\Delta\Lambda_{02}\left(Y_{i 2}\right)\exp\{h_2(\boldsymbol{x}_i)\}\right]^{\left(1-\delta_{i 1}\right) \delta_{i 2}} \times \left[\Delta\Lambda_{03}\left(Y_{i 2} \right)\exp\{h_3(\boldsymbol{x}_i)\}\right]^{\delta_{i 1} \delta_{i 2}} \\[2ex]
    &\times \exp\left\{-\gamma_i\left[\Lambda_{01}\left(Y_{i 1}\right)\exp\{h_1(\boldsymbol{x}_i)\} + \Lambda_{02}\left(Y_{i 1}\right)\exp\{h_2(\boldsymbol{x}_i)\} + \delta_{i 1}[\Lambda_{03}(Y_{i 2}) - \Lambda_{03}(Y_{i 1})] \exp\{h_3(\boldsymbol{x}_i)\}\right]\right\}.
\end{aligned}
\]

\vspace{2ex}

\noindent Considering only those terms which involve $\gamma_i$, we can reduce the above expression to

\begin{align}
\begin{split}
\label{eq:gammapost}
    f&(\boldsymbol{\gamma} \mid \mathcal{D}, \boldsymbol{\psi}) \propto  \prod_{i = 1}^{n} \gamma_i^{\frac{1}{\theta} + \delta_{i 1} + \delta_{i 2} - 1} \times \exp\left\{-\gamma_i \left[\frac{1}{\theta} + \Lambda_{01}\left(Y_{i1}\right) \exp\{h_1(\boldsymbol{x}_i)\} \right.\right. \\
    &\left.\left. +\ \Lambda_{02}\left(Y_{i1}\right) \exp\{h_2(\boldsymbol{x}_i)\} + \delta_{i 1}[\Lambda_{03}(Y_{i2}) - \Lambda_{03}(Y_{i1})] \exp\{h_3(\boldsymbol{x}_i)\}\right]\right\},
\end{split}
\end{align}

\vspace{2ex}

\noindent which is recognized to  be the `kernel' of a Gamma distribution. Thus, conditional on the data, the $\gamma_i$'s follow a $\text{Gamma}(\tilde{a},\ \tilde{b})$ distribution with

\begin{align}
    \tilde{a} &= \frac{1}{\theta} + \delta_{i1} + \delta_{i2} \label{eq:atidle} \\
    \tilde{b} &= \frac{1}{\theta} + \Lambda_{01}\left(Y_{i 1}\right) \exp\{h_1(\boldsymbol{x}_i)\} +\ \Lambda_{02}\left(Y_{i1}\right) \exp\{h_2(\boldsymbol{x}_i)\} \label{eq:btidle} \\
    &\quad \quad +\ \delta_{i1} \left\{\Lambda_{03}(Y_{i 2}) - \Lambda_{03}(Y_{i1})\right\} \exp\{h_3(\boldsymbol{x}_i)\}. \nonumber
\end{align}

\noindent Hence, the posterior means of the $\gamma_i$ are given by $\mathbb{E}[\gamma_i\mid\mathcal{D}, \boldsymbol{\psi}] = \tilde{a}/\tilde{b}$. The posterior means of $\log(\gamma_i)$ can be derived as follows. Without loss of generality, let the rate parameter, $\tilde{b}$, equal 1, as its effect on the logarithm of $\gamma_i$ is a negative linear shift by a factor of $\log(\tilde{b})$. The density of $\gamma_i\sim\text{Gamma}(\tilde{a}, 1)$ is given by

\[
f(\gamma_i\mid\mathcal{D},\boldsymbol{\psi}) = \frac{1}{\Gamma(\tilde{a})}\gamma_i^{\tilde{a} - 1}\exp\{-\gamma_i\}d\gamma_i = \frac{1}{\Gamma(\tilde{a})}\gamma_i^{\tilde{a}}\exp\{-\gamma_i\}\frac{d\gamma_i}{\gamma_i}. 
\]

\vspace{1ex}

\noindent Taking $\gamma_i = \exp\{\log(\gamma_i)\}$ and $d\gamma_i /\gamma_i = d\log(\gamma_i)$, we have

\begin{equation}
\label{eq:loggammapost1}
    f(\log(\gamma_i)\mid\mathcal{D},\boldsymbol{\psi}) = \frac{1}{\Gamma(\tilde{a})}\exp\{\tilde{a}\log(\gamma_i)-\exp\{\log(\gamma_i)\}\}d\log(\gamma_i).
\end{equation}

\noindent As (\ref{eq:loggammapost1}) is a probability density function, it must integrate to unity. Since $\log(\gamma_i)$ has support in $\mathbb{R}$, we have

\[
\Gamma(\tilde{a}) = \int_{\mathbb{R}}\exp\{\tilde{a}\log(\gamma_i)-\exp\{\log(\gamma_i)\}\}d\log(\gamma_i).
\]

\vspace{1ex}

\noindent Differentiating under the integral with respect to $\tilde{a}$, we have that:

\[
\begin{aligned}
\frac{\partial}{\partial\tilde{a}}&\left[\exp\{\tilde{a}\log(\gamma_i)-\exp\{\log(\gamma_i)\}\}d\log(\gamma_i)\right] = \log(\gamma_i)\exp\{\tilde{a}\log(\gamma_i)-\exp\{\log(\gamma_i)\}\}d\log(\gamma_i) \\
&= \Gamma(\tilde{a})\log(\gamma_i)f(\log(\gamma_i)\mid\mathcal{D},\boldsymbol{\psi}).
\end{aligned}
\]

\vspace{1ex}

\noindent Dividing by $\Gamma(\tilde{a})$ and integrating over $\mathbb{R}$ with respect to $\log(\gamma_i)$ yields the posterior expectation, given by

\[
\begin{aligned}
    \mathbb{E}&[\log(\gamma_i) \mid \mathcal{D}, \boldsymbol{\psi}] = -\log(\tilde{b}) + \int_{\mathbb{R}} \log(\gamma_i)f(\log(\gamma_i) \mid \mathcal{D},\boldsymbol{\psi})\\
    &= -\log(\tilde{b}) + \frac{1}{\Gamma(\tilde{a})}\int_{\mathbb{R}} \frac{\partial}{\partial\tilde{a}}\exp\{\tilde{a}\log(\gamma_i)-\exp\{\log(\gamma_i)\}\}d\log(\gamma_i)\\
    &= -\log(\tilde{b}) + \frac{1}{\Gamma(\tilde{a})}\frac{\partial}{\partial\tilde{a}}\int_{\mathbb{R}} \exp\{\tilde{a}\log(\gamma_i)-\exp\{\log(\gamma_i)\}\}d\log(\gamma_i)\\
    &= -\log(\tilde{b}) + \frac{1}{\Gamma(\tilde{a})}\frac{\partial}{\partial\tilde{a}}\Gamma(\tilde{a})= -\log(\tilde{b}) + \frac{\partial}{\partial\tilde{a}}\log\left[\Gamma(\tilde{a})\right]\\
    &= \text{digamma}(\tilde{a}) - \log(\tilde{b}),
\end{aligned}
\]

\vspace{1ex}

\noindent where $\text{digamma}(\tilde{a}) = \partial\log[\Gamma(\tilde{a})]/\partial\tilde{a}$ and $\Gamma(\cdot)$ is the gamma function. 

\subsection{E-Step}
\label{sec:b.2}

The E-step calculates the expected log-conditional likelihood of the `complete' data given the observed data, $\mathcal{D}$, and the current estimates of our parameters, denoted by $\boldsymbol{\psi}_c$. The M-step later will maximize this expectation. However, as the maximizer of this objective function over the space of absolutely continuous cumulative baseline hazards does not exist \citep{li2000covariate}, we restrict the parameter space of the cumulative baseline hazards to the one containing piecewise constant functions, with jumps occurring at observed event times. Maximizers over this discrete space are termed nonparametric maximum likelihood estimates of $\Lambda_{01}, \Lambda_{02}$, and $\Lambda_{03}$. Under this parameter space, $\lambda_{0g}(t)$ in $Q$ is replaced by $\Delta\Lambda_{0g}(t)$, the jump size at $t$ for the baseline hazards of each state transition \citep{li2000covariate}, and $\Lambda_{0g}(t) = \sum_{s = 0}^t \Delta\Lambda_{0g}(s)$. Note that $\Delta\Lambda_{0g}(s) = 0$ if $s$ is not one of the observed event times corresponding to  state transition $g$. Thus we work with $\tilde{L}(\boldsymbol{\psi} ; \mathcal{D}, \boldsymbol{\gamma})$, defined over the new parameter space, and derive 
\[
\begin{aligned}
    Q&\left(\boldsymbol{\psi} \mid \boldsymbol{\psi}_c\right) = \mathbb{E}\left[\log \tilde{L}(\boldsymbol{\psi} ; \mathcal{D}, \boldsymbol{\gamma}) \mid \mathcal{D}, \boldsymbol{\psi}_c\right] \\[2ex]
    &= \mathbb{E}\left[\log\left(\prod_{i = 1}^{n} \gamma_i^{\delta_{i1} + \delta_{i2}} \frac{\theta^{-\frac{1}{\theta}}}{\Gamma\left(\frac{1}{\theta}\right)} \gamma_i^{\frac{1}{\theta}-1} \exp\left\{-\frac{\gamma_i}{\theta}\right\} \times [\Delta\Lambda_{01}(Y_{i1}) \exp\{h_1(\boldsymbol{x}_i)\}]^{\delta_{i1}}\right.\right. \\[1ex]
    &\quad \quad \times [\Delta\Lambda_{02}(Y_{i2}) \exp\{h_2(\boldsymbol{x}_i)\}]^{(1 - \delta_{i1}) \delta_{i2}} \times [\Delta\Lambda_{03}(Y_{i2})\exp\{h_3(\boldsymbol{x}_i)\}]^{\delta_{i1} \delta_{i2}} \\[1ex]
    &\quad \quad \times \exp\{-\gamma_i[\Lambda_{01}(Y_{i1}) \exp\{h_1(\boldsymbol{x}_i)\} + \Lambda_{02}(Y_{i1}) \exp\{h_2(\boldsymbol{x}_i)\} \\[1ex]
    &\quad \quad \quad \quad \left.\left. +\ \delta_{i1} \left\{\Lambda_{03}(Y_{i2}) - \Lambda_{03}(Y_{i1})\right\} \exp\{h_3(\boldsymbol{x}_i)\}]\}\right) \mid \mathcal{D}, \boldsymbol{\psi}_c\right] \\[2ex]
    &= \mathbb{E} \left[\sum_{i = 1}^{n} \delta_{i1}\log(\gamma_i) + \delta_{i2}\log(\gamma_i) + \delta_{i1}\log[\Delta\Lambda_{01}(Y_{i1})] + \delta_{i1}h_1(\boldsymbol{x_i}) + (1 - \delta_{i1})\delta_{i2}\log[\Delta\Lambda_{02}(Y_{i2})] \right. \\[1ex] 
    &\quad \quad +\ (1 - \delta_{i1})\delta_{i2}h_2(\boldsymbol{x_i}) + \delta_{i1}\delta_{i2}\log[\Delta\Lambda_{03}(Y_{i2})] + \delta_{i1}\delta_{i2}h_3(\boldsymbol{x_i}) - \gamma_i[\Lambda_{01}(Y_{i1}) \exp\{h_1(\boldsymbol{x}_i)\} \\[1ex]
    &\quad \quad +\ \Lambda_{02}(Y_{i1}) \exp\{h_2(\boldsymbol{x}_i)\} + \delta_{i1} \left\{\Lambda_{03}(Y_{i2}) - \Lambda_{03}(Y_{i1})\right\} \exp\{h_3(\boldsymbol{x}_i)\}] - \frac{1}{\theta} \log(\theta) \\[1ex]
\end{aligned}
\]

\[
\begin{aligned}
    &\quad \quad \left. +\ \left(\frac{1}{\theta} - 1 \right) \log(\gamma_i) - \frac{1}{\theta} \gamma_i - \log\left[\Gamma\left(\frac{1}{\theta}\right)\right] \mid \mathcal{D}, \boldsymbol{\psi}_c\right] \\[2ex]
    &= \sum_{i = 1}^{n} \delta_{i1} \mathbb{E}\left[\log(\gamma_i) \mid \mathcal{D}, \boldsymbol{\psi}_c\right] + \delta_{i2} \mathbb{E}\left[\log(\gamma_i) \mid \mathcal{D}, \boldsymbol{\psi}_c\right] + \delta_{i1} \log[\Delta\Lambda_{01}(Y_{i1})] + \delta_{i1} h_1(\boldsymbol{x_i}) \\[1ex]
    &\quad \quad +\ (1 - \delta_{i1}) \delta_{i2} \log[\Delta\Lambda_{02}(Y_{i2})] + (1 - \delta_{i1}) \delta_{i2} h_2(\boldsymbol{x_i}) + \delta_{i1} \delta_{i2} \log[\Delta\Lambda_{03}(Y_{i2})] + \delta_{i1} \delta_{i2} h_3(\boldsymbol{x_i}) \\[1ex]
    &\quad \quad -\ \mathbb{E}\left[\gamma_i \mid \mathcal{D}, \boldsymbol{\psi}_c\right] \left[\Lambda_{01}(Y_{i1}) \exp\{h_1(\boldsymbol{x}_i)\} + \Lambda_{02}(Y_{i1}) \exp\{h_2(\boldsymbol{x}_i)\}\right. \\[1ex]
    &\quad \quad \quad \quad \left.+\ \delta_{i1} \left\{\Lambda_{03}(Y_{i2}) - \Lambda_{03}(Y_{i1})\right\} \exp\{h_3(\boldsymbol{x}_i)\}\right] \\[1ex]
    &\quad \quad -\ \frac{1}{\theta} \log(\theta) + \left(\frac{1}{\theta} - 1\right) \mathbb{E}\left[\log(\gamma_i) \mid \mathcal{D}, \boldsymbol{\psi}_c\right] -\frac{1}{\theta}\mathbb{E}\left[\gamma_i \mid \mathcal{D}, \boldsymbol{\psi}_c\right] -\log\left[\Gamma\left(\frac{1}{\theta}\right)\right] \\[2ex]
    &= Q_1 + Q_2 + Q_3 + Q_4,
\end{aligned}
\]

\vspace{1ex}

\noindent where $Q_1$, $Q_2$, $Q_3$, and $Q_4$ are additive pieces of `$Q$', each involving non-overlapping unknown parameters:

\[
\begin{aligned}
    Q_1 &= \sum_{i = 1}^{n} \delta_{i1} \mathbb{E}\left[\log(\gamma_i) \mid \mathcal{D}, \boldsymbol{\psi}_c\right] + \delta_{i1} \left\{\log\left[\Delta\Lambda_{01}\left(Y_{i 1}\right)\right] + h_1(\boldsymbol{x_i})\right\} \\
    &\quad\quad -\ \mathbb{E}\left[\gamma_i \mid \mathcal{D}, \boldsymbol{\psi}_c\right] \Lambda_{01}\left(Y_{i1}\right) \exp\{h_1(\boldsymbol{x}_i)\} \\[2ex]
    Q_2 &= \sum_{i = 1}^{n} \delta_{i2} \mathbb{E} \left[\log(\gamma_i) \mid \mathcal{D}, \boldsymbol{\psi}_c\right] + \left(1 - \delta_{i1}\right) \delta_{i2}\left\{ \log\left[\Delta\Lambda_{02}\left(Y_{i2}\right)\right] + h_2(\boldsymbol{x_i})\right\}\\
    &\quad\quad -\ \mathbb{E}\left[\gamma_i \mid \mathcal{D}, \boldsymbol{\psi}_c\right] \Lambda_{02}\left(Y_{i1}\right) \exp\{h_2(\boldsymbol{x}_i)\} \\
    Q_3 &= \sum_{i = 1}^{n} \delta_{i1} \delta_{i2} \left\{\log\left[\Delta\Lambda_{03}\left(Y_{i2}\right)\right] + h_3(\boldsymbol{x_i})\right\} \\
    &\quad\quad -\ \mathbb{E}\left[\gamma_i \mid \mathcal{D}, \boldsymbol{\psi}_c\right] \delta_{i1} \left\{\Lambda_{03}\left(Y_{i2}\right) - \Lambda_{03}\left(Y_{i1}\right)\right\} \exp\{h_3(\boldsymbol{x}_i)\} \\[2ex]
    Q_4 &= \sum_{i = 1}^{n} -\frac{1}{\theta} \log(\theta) + \left(\frac{1}{\theta} - 1\right) \mathbb{E}\left[\log(\gamma_i) \mid \mathcal{D}, \boldsymbol{\psi}_c\right] - \frac{1}{\theta} \mathbb{E}\left[\gamma_i \mid \mathcal{D}, \boldsymbol{\psi}_c\right] - \log\left[\Gamma\left(\frac{1}{\theta}\right)\right].
\end{aligned}
\]

\subsection{M-Step}
\label{sec:b.3}

  As shown in the previous section, our objective function, $Q(\boldsymbol{\psi} \mid \mathcal{D}, \boldsymbol{\psi}_c)$, can be written as the sum of pieces $Q_1$, $Q_2$, $Q_3$, and $Q_4$. Each of the first three involves only the baseline hazard and $h$ functions for a state transition, and the last one involves only the frailty variance. Thus, the M-step updates for $\Lambda_{0g}, h_{g}, g=1,2,3$,   can be defined utilizing $Q_1$, $Q_2$, and $Q_3$, separately, and the frailty variance, $\theta$, with $Q_4$.   Finally, because both $\Lambda_{0g}, h_g$ are nonparametric, we adopt a profiling approach to facilitate maximization. Specifically, for each $g=1,2,3$, we maximize $Q_g$ with respect to the jump sizes of $\Lambda_{0g}$, while fixing  $h_g$, and obtain the nonparametric  estimates as follows. 

\vspace{1ex}

%As such, the M-step updates for these jump sizes, $\Delta\Lambda_{0g}(t)$, can be derived as follows.\todo{Come back to this section and align with new notation}

\noindent First,  for any fixed $t>0$, differentiating $Q_1$ with respect to $\Delta\Lambda_{01}(t)$, the jump size at $t$, the score function for $\Delta\Lambda_{01}(t)$ is 
\[
\frac{\partial Q_1}{\partial \Delta\Lambda_{01}(t)} = \sum_{i = 1}^{n}\frac{\delta_{i1} I(Y_{i1} = t)}{\Delta\Lambda_{01}(t)} - \mathbb{E}\left[\gamma_i \mid \mathcal{D}, \boldsymbol{\psi}_c\right] I\left(Y_{i1} \geq t\right)\exp\{h_1(\boldsymbol{x}_i)\}.
\]
%where by convention, we define $0/0=0$.
\vspace{1ex}

\noindent Setting this equal to zero,  the maximizer, $\widehat{\Delta\Lambda_{01}}(t)$ is obtained as 

\begin{equation}
\label{eq:m1}
 \widehat{\Delta\Lambda_{01}}(t)  = \frac{\sum_{i=1}^{n} \delta_{i1} I\left(Y_{i1} = t\right)
}{\sum_{i=1}^{n}\mathbb{E}\left[\gamma_i\mid\mathcal{D}, \boldsymbol{\psi}_c\right] I\left(Y_{i1} \geq t\right) \exp\left\{h_{1,c}(\boldsymbol{x}_i)\right\}}.   
\end{equation}

\vspace{1ex}

%\noindent where the numerator is the observed number of non-terminal events.

%\vspace{6ex}

\noindent Then, differentiating $Q_2$ with respect to $\Delta\Lambda_{02}(t)$, we have the score function

\[
\frac{\partial Q_2}{\partial \Delta\Lambda_{02}(t)} = \sum_{i = 1}^{n} \frac{\left(1 - \delta_{i1}\right) \delta_{i2}  I\left(Y_{i2} = t\right)}{\Delta\Lambda_{02}(t)} - \mathbb{E}\left[\gamma_i \mid \mathcal{D}, \boldsymbol{\psi}_c\right] I\left(Y_{i1} \geq t\right) \exp\{h_2(\boldsymbol{x}_i)\}.
\]

\vspace{1ex}

\noindent By setting this to zero and solving for $\Delta\Lambda_{02}(t)$, we obtain the maximizer, 
$\widehat{\Delta\Lambda_{02}}(t)$, as
\begin{equation}
\label{eq:m2}
  %  \Delta\Lambda_{02}^{(m + 1)}(t) 
    \widehat{\Delta\Lambda_{02}}(t)
    = \frac{\sum_{i = 1}^{n}\left(1 - \delta_{i1}\right) \delta_{i2} I\left(Y_{i2} = t\right)}{\sum_{i = 1}^{n} \mathbb{E}\left[\gamma_i \mid \mathcal{D}, \boldsymbol{\psi}_c\right] I\left(Y_{i1} \geq t\right)\exp\left\{h_{2,c}(\boldsymbol{x}_i)\right\}}.
\end{equation}

\vspace{1ex}

%\noindent where the numerator is the number of terminal events observed prior to non-terminal events.

\noindent Finally, differentiating $Q_3$ with respect to $\Delta\Lambda_{03}(t)$ yields a score function for $\Delta\Lambda_{03}(t)$:

\[
\frac{\partial Q_3}{\partial \Delta\Lambda_{03}(t)} = \sum_{i = 1}^{n}\frac{\delta_{i1} \delta_{i2} I\left(Y_{i2} = t\right)}{\Delta\Lambda_{03}\left(t\right)} - \mathbb{E}\left[\gamma_i \mid  \mathcal{D}, \boldsymbol{\psi}_c\right] \delta_{i1} \left[I(Y_{i2} \geq t) - I(Y_{i1} \geq t)\right]\exp\{h_3(\boldsymbol{x}_i)\}.
\]

\vspace{1ex}

\noindent Equating this to zero and solving it, we have the maximizer, $\widehat{\Delta\Lambda_{03}}(t)$,  as:

\begin{equation}
\label{eq:m3}
 %   \Delta\Lambda_{03}^{(m + 1)}(t) 
    \widehat{\Delta\Lambda_{03}}(t)
    = \frac{\sum_{i = 1}^{n} \delta_{i1} \delta_{i2} I\left(Y_{i2} = t\right)}{\sum_{i = 1}^{n} \mathbb{E}\left[\gamma_i \mid  \mathcal{D}, \boldsymbol{\psi}_c\right] \delta_{i1} \left[I(Y_{i2} \geq t) - I(Y_{i1} \geq t)\right] \exp\left\{h_{3,c}(\boldsymbol{x}_i)\right\}}.
\end{equation}
\vspace{1ex}

%\noindent where the numerator reflects the number of terminal events observed after non-terminal events. Note that these closed form updates in the M-step are Breslow-type estimators. 
%As such, to seed the EM algorithm, we initialize $\Lambda_{01}, \Lambda_{02}$, and $\Lambda_{03}$ with their respective, unadjusted Nelson-Aalen estimators.

\subsection{N-Step}
%\label{sec:b.3}

Plugging the estimates, i.e., $\widehat{\Delta\Lambda_{0g}}(t),$ into $Q_g (g=1,2,3)$,  (by noting that
$\widehat{\Lambda_{0g}}(t)= \sum_{s \le t}
\widehat{\Delta\Lambda_{0g}}(s)$)
yields 
the expected  log-profile likelihoods for $h_g, g=1,2,3$.   That is, with an added subscript $P$ (for profile), we have that  

\noindent 
%$\mathbb{E}_{\boldsymbol{\gamma}}\left[\ell\left(\boldsymbol{\psi}; \mathcal{D}, \boldsymbol{\gamma}\right)\mid \mathcal{D}, \boldsymbol{\psi}\right] = 
%$Q_{1, P}= \sum_{i = 1}^{n} \delta_{i1} \mathbb{E}\left[\log(\gamma_i) \mid \mathcal{D}, \boldsymbol{\psi}\right] + \delta_{i2} \mathbb{E}\left[\log(\gamma_i) \mid \mathcal{D}, \boldsymbol{\psi}\right]$

\vspace{2ex}

 $Q_{1, P} = \sum_{i = 1}^{n} \delta_{i1} \left\{h_1(\boldsymbol{x_i}) - \log[\sum_{j=1}^{n}\mathbb{E}\left[\gamma_j\mid\mathcal{D}, \boldsymbol{\psi}_c\right] I\left(Y_{j1} \geq  Y_{i1}\right) \exp\left\{h_1(\boldsymbol{x}_j)\right\}]\right\} - |D_1| + \sum_{i=1}^{n}\delta_{i 1} \mathbb{E}[\log(\gamma_i) | \mathcal{D}, \boldsymbol{\psi}_c],$

\vspace{2ex}

$ Q_{2, P} = \sum_{i = 1}^{n}  (1 - \delta_{i1}) \delta_{i2}\left\{h_2(\boldsymbol{x_i}) - \log[\sum_{j = 1}^{n} \mathbb{E}\left[\gamma_j \mid \mathcal{D}, \boldsymbol{\psi}_c\right] I\left(Y_{j2} \geq Y_{i2}\right)\exp\left\{h_2(\boldsymbol{x}_j)\right\}] \right\}- |D_2|+ \sum_{i=1}^{n}\delta_{i 2} \mathbb{E}[\log(\gamma_i) | \mathcal{D}, \boldsymbol{\psi}_c],$

\vspace{2ex}

$ Q_{3, P} = \sum_{i = 1}^{n} \delta_{i1} \delta_{i2}\left\{h_3(\boldsymbol{x_i}) - \log[\sum_{j = 1}^{n} \mathbb{E}\left[\gamma_j \mid  \mathcal{D}, \boldsymbol{\psi}_c\right] \delta_{j1} I\left[Y_{j2} \geq \max(Y_{i2}, Y_{j1})\right] \exp\left\{h_3(\boldsymbol{x}_j)\right\}]\right\}- |D_3|$,

\vspace{2ex}

\noindent where $|D_1|, |D_2|, |D_3|$ are the numbers of the observed progressions, deaths prior to progression, and deaths following progression, respectively.  Note that in each of $Q_{1,P}, Q_{2,P}, Q_{3,P}$, only the first term involves $h_g$ that needs to be maximized with respect to. To avoid redundancy, in the main text, we use the notations $Q_{1,P}, Q_{2,P}, Q_{3,P}$ to specifically refer to the first term in each expression, as the remaining terms can be treated as constant in the context of maximization with respect to $h_g$.
The detailed derivations follow those in a standard Cox model setting  \cite{johansen1983extension}.

 %--- APPENDIX C -----------------------------------------------------------------

%\newpage

\setcounter{figure}{0}
\setcounter{table}{0}
\setcounter{equation}{0}

\section{Bivariate Brier Score}
\label{sec:c}

We provide additional details of the derivation of the bivariate Brier score outlined in Section 4 in the main text. Namely, we show that the expectation of the bivariate Brier score is equal to the mean squared error of the predictor, $\pi_i(t)$, plus a constant. To proceed, we compute the expectation in additive pieces. In the first piece, we consider the region where at least the non-terminal event is observed by time $t$, and $Y_{i1}$ is less than or equal to $Y_{i2}$, but the terminal event may or may not be observed. In the second piece, we consider the region where the terminal event is observed prior to the non-terminal event occurring. In the third piece, we consider the region where neither event has been observed by time $t$.

\vspace{4ex}

\noindent \textbf{Piece 1}: At least the non-terminal event is observed by time $t$, and $Y_{i1}$ is less than or equal to $Y_{i2}$, but the terminal event may or may not be observed.

\begin{align}
    \mathbb{E}&\left[\frac{\pi_i(t)^2 \times I\left(Y_{i1} \leq t,\ \delta_{i1} = 1,\ Y_{i1} \leq Y_{i2}\right)}{G_i(Y_{i1})}\right] = \pi_i(t)^2 \times\mathbb{E}\left[\frac{I\left(T_{i1} \leq t,\ T_{i1} \leq C_i,\ T_{i1} \leq T_{i2}\right)}{G_i(T_{i1})}\right] \nonumber \\[2ex]
    &= \pi_i(t)^2 \times\mathbb{E}\left\{\mathbb{E}\left[\frac{I\left(T_{i1} \leq t,\ T_{i1} \leq C_i,\ T_{i1} \leq T_{i2}\right)}{G_i(Y_{i1})}\mid T_{i1},T_{i2}\right]\right\} \nonumber \\[2ex]
    &= \pi_i(t)^2 \times\mathbb{E}\left\{\frac{I\left(T_{i1} \leq t,\  T_{i1} \leq T_{i2}\right)}{G_i(T_{i1})} \times \mathbb{E}\left[I\left(T_{i1} \leq C_i\right)\mid T_{i1},T_{i2}\right]\right\} \nonumber \\[2ex]
    &= \pi_i(t)^2 \times\mathbb{E}\left[\frac{I\left(T_{i1} \leq t,\  T_{i1} \leq T_{i2}\right)}{G_i(T_{i1})} \times \Pr\left(T_{i1} \leq C_i\right)\right] \nonumber \\[2ex]
    &= \pi_i(t)^2 \times\mathbb{E}\left[\frac{I\left(T_{i1} \leq t,\  T_{i1} \leq T_{i2}\right)}{G_i(T_{i1})} \times G_i\left(T_{i1}\right)\right] \nonumber \\[2ex]
    &= \pi_i(t)^2 \times\mathbb{E}\left[I\left(T_{i1} \leq t,\  T_{i1} \leq T_{i2}\right)\right] = \pi_i(t)^2 \times\Pr\left(T_{i1} \leq t,\  T_{i1} \leq T_{i2}\right). \label{eq:bbs1}
\end{align}

\vspace{4ex}

\noindent \textbf{Piece 2}: The terminal event is observed prior to the non-terminal event occurring. 

\begin{align}
    \mathbb{E}&\left[\frac{\pi_i(t)^2 \times I\left(Y_{i1} \leq t,\ Y_{i2} \leq t,\ \delta_{i1} = 0,\ \delta_{i2} = 1,\ Y_{i1} \leq Y_{i2}\right)}{G_i(Y_{i2})}\right] \nonumber \\[2ex]
    &= \pi_i(t)^2 \times\mathbb{E}\left[\frac{I\left(T_{i2} \leq t,\ T_{i1} > T_{i2},\ T_{i2} \leq C_i\right)}{G_i(Y_{i2})}\right] \nonumber \\[2ex]
    &= \pi_i(t)^2 \times\mathbb{E}\left\{\mathbb{E}\left[\frac{I\left(T_{i2} \leq t,\ T_{i1} > T_{i2},\ T_{i2} \leq C_i\right)}{G_i(Y_{i2})}\mid T_{i1}, T_{i2}\right]\right\}\nonumber \\[2ex]
    &= \pi_i(t)^2 \times\mathbb{E}\left\{\frac{I\left(T_{i2} \leq t,\ T_{i1} > T_{i2}\right)}{G_i(Y_{i2})} \times \mathbb{E}\left[I\left(T_{i2} \leq C_i\right)\mid T_{i1}, T_{i2}\right]\right\} \nonumber \\[2ex]
    &= \pi_i(t)^2 \times\mathbb{E}\left[\frac{I\left(T_{i2} \leq t,\ T_{i1} > T_{i2}\right)}{G_i(Y_{i2})} \times \Pr\left(T_{i2} \leq C_i\right)\right] \nonumber \\[2ex]
    &= \pi_i(t)^2 \times\mathbb{E}\left[\frac{I\left(T_{i2} \leq t,\ T_{i1} > T_{i2}\right)}{G_i(Y_{i2})} \times G_i\left(T_{i2}\right)\right] \nonumber \\[2ex]
    &= \pi_i(t)^2 \times\mathbb{E}\left[I\left(T_{i2} \leq t,\ T_{i1} > T_{i2}\right)\right] = \pi_i(t)^2 \times\Pr\left(T_{i2} \leq t,\ T_{i1} > T_{i2}\right). \label{eq:bbs2}
\end{align}

\vspace{4ex}

\noindent \textbf{Piece 3}: Neither event has been observed by time $t$.

\begin{align}
    \mathbb{E}&\left[\frac{[1 - \pi_i(t)]^2 \times I\left(Y_{i1} > t,\ Y_{i2} > t\right)}{G_i(t)}\right] = \frac{[1 - \pi_i(t)]^2}{G_i(t)} \times\mathbb{E}\left[I\left(T_{i1} > t,\ T_{i2} > t,\ C_i > t\right)\right] \nonumber \\[2ex]
    &= \frac{[1 - \pi_i(t)]^2}{G_i(t)} \times\Pr\left(T_{i1} > t,\ T_{i2} > t,\ C_i > t\right) \nonumber \\[2ex]
    &= \frac{[1 - \pi_i(t)]^2}{G_i(t)} \times\Pr\left(T_{i1} > t,\ T_{i2} > t\right) \times \Pr\left(C_i > t\right) \quad \text{since}\quad T_{i1},T_{i2}\perp C_i \nonumber \\[2ex]
    &= \frac{[1 - \pi_i(t)]^2}{G_i(t)} \times S_i(t) \times G_i\left(t\right) = [1 - \pi_i(t)]^2 \times S_i(t). \label{eq:bbs3}
\end{align}

\vspace{2ex}

\noindent Combining (\ref{eq:bbs1}) - (\ref{eq:bbs3}), and summing over the $n$ individuals, we can see that

\[
\begin{aligned}
    \mathbb{E}&\left[{\rm BBS}(t)\right] = \mathbb{E}\left[\frac{1}{n}\sum_{i = 1}^n\frac{\pi_i(t)^2\cdot I\left(Y_{i1} \leq t,\ \delta_{i1} = 1,\ Y_{i1} \leq Y_{i2}\right)}{G_i(Y_{i1})}\right.\\
    &\quad \left. +\ \frac{\pi_i(t)^2\cdot I\left(Y_{i1} \leq t,\ Y_{i2} \leq t,\ \delta_{i1} = 0,\ \delta_{i2} = 1,\ Y_{i1} \leq Y_{i2}\right)}{G_i(Y_{i2})} + \frac{[1 - \pi_i(t)]^2\cdot I\left(Y_{i1} > t,\ Y_{i2} > t\right)}{G_i(t)}\right]\\
    &= \frac{1}{n}\sum_{i=1}^n\mathbb{E}\left[\frac{\pi_i(t)^2\cdot I\left(Y_{i1} \leq t,\ \delta_{i1} = 1,\ Y_{i1} \leq Y_{i2}\right)}{G_i(Y_{i1})}\right]\\
    &\quad +\ \mathbb{E}\left[\frac{\pi_i(t)^2\cdot I\left(Y_{i1} \leq t,\ Y_{i2} \leq t,\ \delta_{i1} = 0,\ \delta_{i2} = 1,\ Y_{i1} \leq Y_{i2}\right)}{G_i(Y_{i2})}\right] + \mathbb{E}\left[\frac{[1 - \pi_i(t)]^2\cdot I\left(Y_{i1} > t,\ Y_{i2} > t\right)}{G_i(t)}\right]\\
    &= \frac{1}{n}\sum_{i=1}^n \pi_i(t)^2\cdot\Pr\left(T_{i1} \leq t,\  T_{i1} \leq T_{i2}\right) +\ \pi_i(t)^2\cdot\Pr\left(T_{i2} \leq t,\ T_{i1} > T_{i2}\right) + [1 - \pi_i(t)]^2\cdot  S_i(t)\\
    &= \frac{1}{n}\sum_{i=1}^n \pi_i(t)^2\cdot \left[1 - S_i(t)\right] + [1 - \pi_i(t)]^2\cdot S_i(t) \\
    &= \text{MSE}(t) + \frac{1}{n}\sum_{i=1}^n S_i\left(t\right)\cdot \left[1 - S_i\left(t\right)\right].
\end{aligned}
\]

\vspace{1ex}

\noindent In expectation, the bivariate Brier score is equivalent to the mean squared error of the predictor plus an additional piece that is constant with respect to $\pi_i(t)$. This term represents the irreducible error incurred by approximating $S_i(t)$.
%b step functions, $I\left(Y_{i1} > t, Y_{i2} > t\right)$.

\end{appendix}

\end{document}